\documentclass[10pt,twocolumn,letterpaper]{article}

\usepackage[pagenumbers]{cvpr} % To force page numbers, e.g. for an arXiv version

\newcommand{\para}[1]{\vspace{.5em}\noindent\textbf{#1.}}

\definecolor{cvprblue}{rgb}{0.21,0.49,0.74}
\usepackage[pagebackref,breaklinks,colorlinks,allcolors=cvprblue]{hyperref}

\usepackage{tabularx}
\usepackage{graphicx}
\usepackage{booktabs}
\usepackage{multirow} 
\usepackage[dvipsnames]{xcolor}
\usepackage{array}
\newcolumntype{?}{!{\vrule width 0.7pt}}
\usepackage{makecell}
\usepackage[utf8]{inputenc}
\usepackage{todonotes}

\def\paperID{0000} % *** Enter the Paper ID here
\def\confName{CVPR}
\def\confYear{2026}

\title{DuoMo: Dual Motion Diffusion for World-Space Human Reconstruction}

\author{
	\fontsize{12pt}{\baselineskip}\selectfont
        Yufu Wang\textsuperscript{1,2} \quad 
        Evonne Ng\textsuperscript{1} \quad 
        Soyong Shin\textsuperscript{1,3} \quad  
        Rawal Khirodkar\textsuperscript{1} \quad  
        Yuan Dong\textsuperscript{1} \quad 
        Zhaoen Su\textsuperscript{1} \\
        \fontsize{12pt}{\baselineskip}\selectfont
        Jinhyung Park\textsuperscript{1,3} \quad
        Kris Kitani\textsuperscript{1,3} \quad  
        Alexander Richard\textsuperscript{1} \quad 
        Fabian Prada\textsuperscript{1} \quad 
        Michael Zollhöfer\textsuperscript{1}\\[0.5em]
        \fontsize{12pt}{\baselineskip}\selectfont
        $^1$ Meta Reality Labs ~
        $^2$ University of Pennsylvania ~
        $^3$ Carnegie Mellon University ~
    }

\begin{document}
% \maketitle
\twocolumn[\maketitle\vspace{-2mm}\begin{center}
    \vspace{-2em}
    \includegraphics[width=0.98\linewidth]{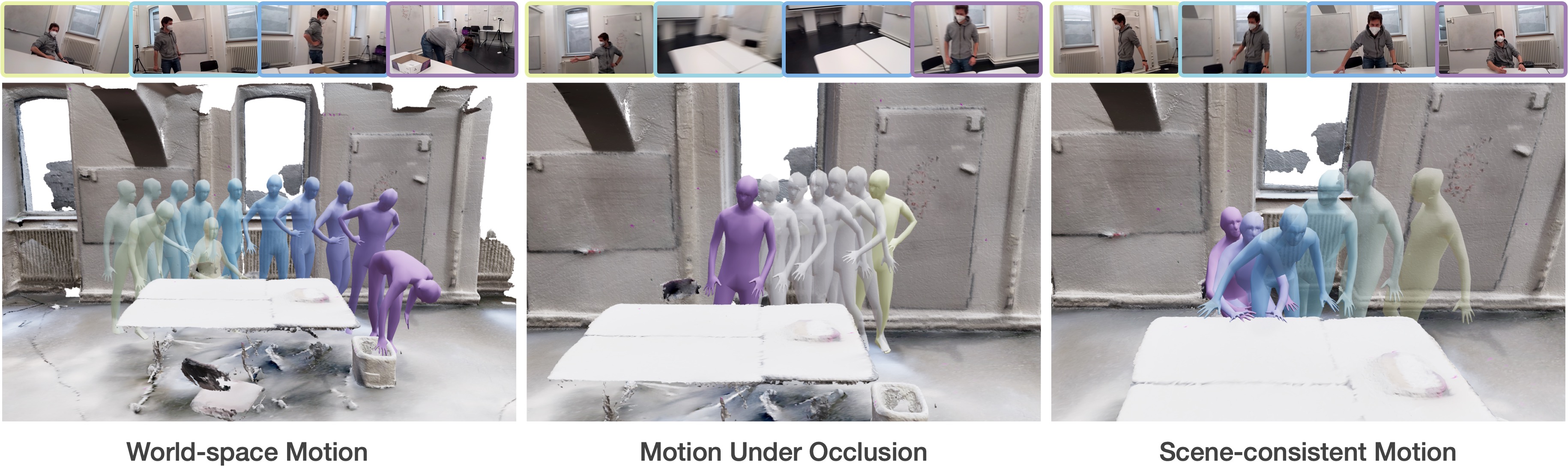}
    \vspace{-2mm}
    \captionof{figure}{\textbf{Dual Motion Diffusion (DuoMo)} recovers world-space human motion from unconstrained monocular videos. Our two-prior approach reconstructs accurate world-space motion (left), completes motion under occlusion or missing observation (middle), and generates scene-consistent motion with guided sampling (right). Moreover, our method outputs the mesh vertices directly without a SMPL model.}
    \label{fig:teaser}
\end{center}\vspace{2mm}\bigbreak]
\begin{abstract}
We present DuoMo, a generative method that recovers human motion in world-space coordinates from unconstrained videos with noisy or incomplete observations. Reconstructing such motion requires solving a fundamental trade-off: \textbf{generalizing} from diverse and noisy video inputs while maintaining \textbf{global motion consistency}. Our approach addresses this problem by factorizing motion learning into two diffusion models. The camera-space model first estimates motion from videos in camera coordinates. The world-space model then lifts this initial estimate into world coordinates and refines it to be globally consistent. Together, the two models can reconstruct motion across diverse scenes and trajectories, even from highly noisy or incomplete observations. Moreover, our formulation is general, generating the motion of mesh vertices directly and bypassing parametric models. DuoMo achieves state-of-the-art performance. On EMDB, our method obtains a 16\% reduction in world-space reconstruction error while maintaining low foot skating. On RICH, it obtains a 30\% reduction in world-space error. Project page: \href{https://yufu-wang.github.io/duomo/}{yufu-wang.github.io/duomo/}
\end{abstract}

\section{Introduction}
\label{sec:intro}
Perceiving human motion within the 3D world is a fundamental aspect of understanding human behavior~\cite{blake2007perception, mcconnell1974understanding}. When viewing a video, humans effortlessly build a coherent mental model of the scene~\cite{peng2018sfv}: we distinguish the subject's movement from the camera's motion, track their trajectories through the environment, and infer their interaction with the world. This perception involves seamlessly integrating a local, view-dependent understanding of a person's pose with a global, persistent model of the world and the motion in it. Replicating this capability in computer vision, however, remains a significant challenge, especially when dealing with unconstrained videos captured by moving cameras.

The focus of human motion modeling in computer vision has recently shifted from analyzing isolated pose sequences to recovering motion grounded in a consistent world coordinate system. This advancement is supported by progress in camera motion estimation and world-space motion capture data. However, current methods face a fundamental trade-off between generalizability and global consistency. On one hand, end-to-end models~\cite{wham, shen2024gvhmr, li2025genmo} can learn powerful priors to predict human motion in world coordinates. However, they often fail to generalize in complex scenes, as they are limited by the motion diversity of the studio-captured data. On the other hand, approaches that first model pose in the camera coordinates and then ``lift" the motion into the world using camera parameters are more robust to diverse motion in the wild~\cite{wang2024tram, wang2025prompthmr}. However, their motion priors are inherently local and often too weak to enforce physical plausibility and consistency in world coordinates.

To address this trade-off, we introduce Dual Motion Diffusion (DuoMo), a method that factors the problem into a two-stage generative process. Instead of committing to a single representation, our approach leverages the strengths of both camera-space and world-space motion modeling by training two independent generative models: 
\begin{itemize}[label=---]
\item {\it a model for camera-space motion} learns a generalizable prior that reconstructs human motion from the camera's perspective, and
\item {\it a model for world-space motion} learns a prior of plausible and consistent human motion in a global coordinate system.
\end{itemize}

Our key insight lies in how we connect these two models. Instead of forcing a single, end-to-end model to learn this complex geometric relationship, we inject this known geometric principle directly into the pipeline. Specifically, the camera-space model's output is explicitly lifted into the world coordinates using the estimated camera motion.

This design choice allows us to effectively structure our system as a two-stage generative process. The camera-space model acts as the initial estimation stage, focusing on interpreting video frames and reconstructing motion with respect to the viewpoint. The explicit lifting step transforms its output into world coordinates, which become a noisy proposal for the world-space motion. All sources of error — from camera pose estimates, depth ambiguity, and the first model's own imperfections — are exposed as noise in the proposal. The world-space model then functions as a global refinement stage, whose task is to take this noisy, incomplete proposal and refine it to be physically plausible and globally consistent.

Our second insight lies in how we define the coordinate system for our world-space model. Many existing methods~\cite{humor, mdm, zhang2024rohm} are trained to operate directly in the canonical coordinate system provided by studio capture datasets (e.g., a lab space where the Y-axis is `up' and the ground is flat at zero height). To use such a model, a reconstruction system must align its camera-space human motion with this canonical space. This alignment requires estimating the camera's pose relative to the scene's ground plane—a step that is ambiguous, error-prone, and often impossible for in-the-wild videos on varied terrain like hills or stairs.

We sidestep this alignment problem entirely. DuoMo's world-space model is not tied to a fixed, canonical coordinate system. Instead, we define the world coordinates for each video relative to its starting camera pose. Our world-space model is then trained to denoise motion directly within these diverse, per-video coordinate systems, without any canonical transformation. This approach is not just simpler; it allows our model to reconstruct plausible motion in complex, in-the-wild scenes.

Finally, our approach is general with respect to the motion representation. Rather than learning to generate the low-dimensional parameters of a body model like SMPL~\cite{loper2023smpl}, our diffusion models generate the motion of the mesh vertices themselves. This shows that our architecture can model geometric motion directly, suggesting a more general path for modeling motion in other categories~\cite{nlf}.

We conduct experiments on EMDB~\cite{kaufmann2023emdb}, RICH~\cite{huang2022capturing} and Egobody~\cite{zhang2022egobody}, and demonstrate substantial improvement in world-space motion accuracy and robustness. To summarize, our main contributions are:
\begin{itemize}%[label=---]
\itemsep0em
  \item DuoMo, a two-stage diffusion method that decouples human motion reconstruction into a camera-space estimation stage and a world-space refinement stage;
  \item a world-space motion model trained to denoise motion in per-video coordinate systems, making it robust to in-the-wild scenes; and
  \item an architecture that generates the motion of mesh vertices directly, without relying on parametric body models.
\end{itemize}

\section{Related work}
\label{sec:related}

\para{Human motion reconstruction} Reconstructing 3D human pose and shape from monocular images~\cite{opt_smplify, hmr, spin, pare, bev, hybrik, pymaf, cliff} and videos~\cite{opt_video, opt_video2, total_capture, vibe, tcmr} has been extensively studied in computer vision. 
Recent advances in transformer architectures~\cite{transformer, vit}, large-scale datasets~\cite{bedlam, amass}, and parametric body models~\cite{smpl, smplx, ghum} have enabled state-of-the-art methods~\cite{hmr2, multihmr, wang2025prompthmr, camerahmr, nlf} to achieve high accuracy on challenging benchmarks~\cite{rich, 3dpw}.

While these methods excel at local pose and shape estimation, reconstructing motion in a consistent world coordinate system remains a significant challenge and has become a recent focus~\cite{glamr, d&d, hps, egobody, emdb}. Current methods for this task broadly fall into two categories: lifting-based and direct prediction. The ``lifting" approach first reconstructs humans in the camera's coordinate system~\cite{slahmr, wang2024tram, wang2025prompthmr, chen2025human3r} and then transform them to world coordinates using estimated cameras~\cite{droid, teed2024dpvo}, often followed by post-hoc refinement steps~\cite{slahmr, josh3r, vidmimic}. In contrast, the ``direct prediction" approach trains a network to regress the human motion in a global coordinate~\cite{trace, wham, shen2024gvhmr, li2025genmo}.
Our method begins with the "lifting" approach, employing a camera-space motion diffusion model as an initial step. Unlike previous methods that rely on post-hoc optimization, we introduce a world-space motion diffusion model to refine the final predictions in world space.

\para{Generative models for motion reconstruction} Generative models~\cite{vae, vqvae, ddpm} have shown great success in different human motion generation tasks~\cite{mdm, zhang2024motiondiffuse, chen2023executing, liu2024emage, li2021ai, hassan2021stochastic}. Applying them for reconstruction tasks presents unique challenges. One line of work employs pre-trained motion priors within an optimization-based framework~\cite{humor, scorehmr, posendf, lu2025dposer}. However, a significant drawback is the computational cost~\cite{humor, slahmr, ho2025phd}.
For faster inference, other work trains conditional generative models to estimate pose or motion~\cite{prohmr, tokenhmr, saleem2025genhmr} based on visual inputs, demonstrating potential to resolve visual ambiguities~\cite{gao2025disrt}. Recent works extend this direction to generate human motion in a consistent global coordinate system~\cite{li2025genmo, zhang2024rohm}. 
Our work reconstructs global human motion with generative models but factors into two priors to gain robustness and generalization.

%\clearpage
\begin{figure*}[h!]
    \centering
      \includegraphics[width=0.99\textwidth]{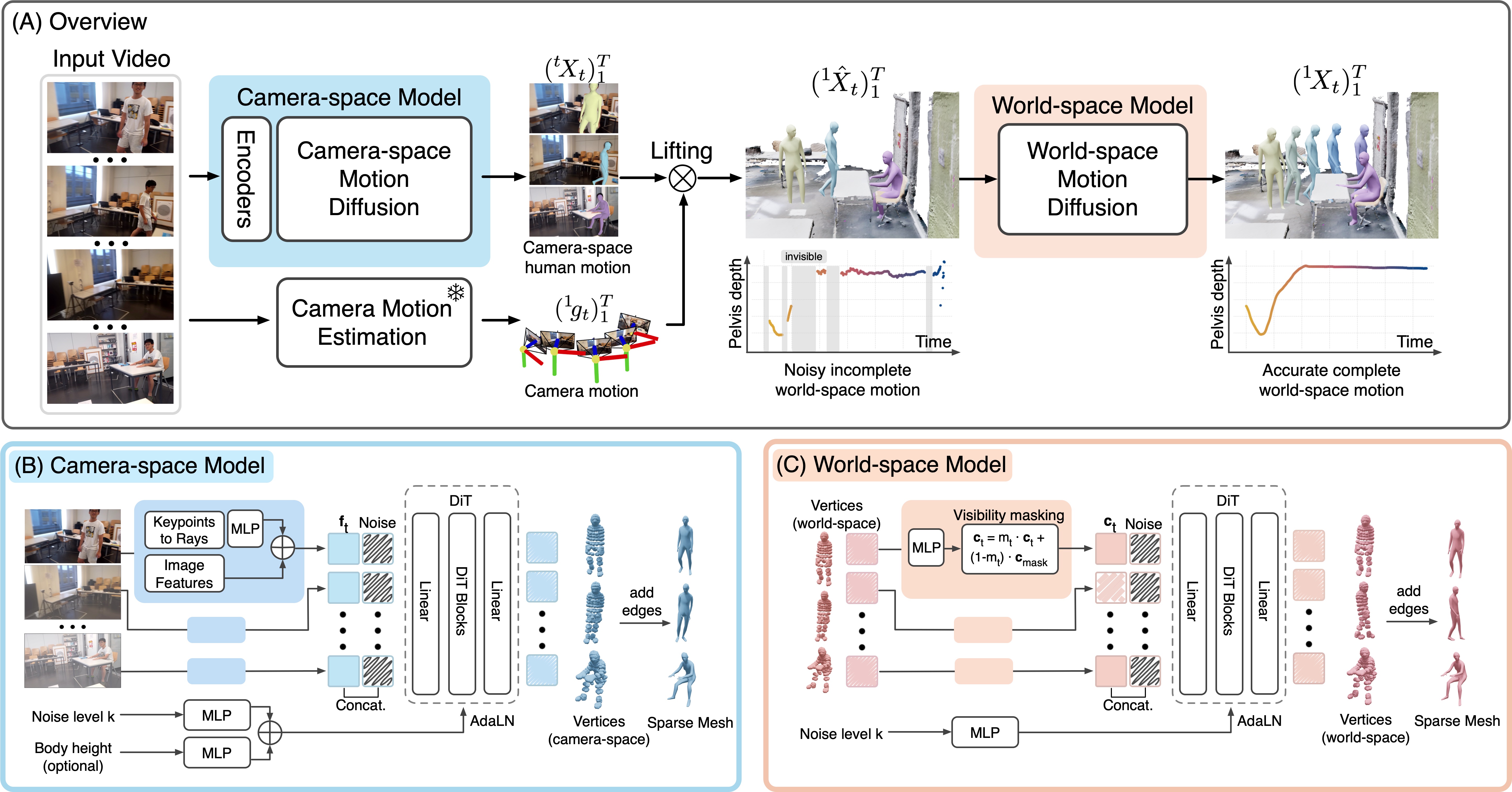}
    \vspace{-2mm}
    \caption{\textbf{Method overview.} (A) In the first stage, our camera-space model encodes video features and generates camera-space human motion. This motion is lifted to the world coordinates using estimated camera poses, becoming the initial proposal for world-space human motion. Some predictions are missing due to subject out of frame. In the second stage, the world-space model encodes the noisy world-space motion and generates globally consistent world-space motion. Plots at the bottom visualize the pelvis depth in the world coordinates. (B) Camera-space model architecture. (C) World-space model architecture. 
    }
    \label{fig:method}
    \vspace{-2mm}
\end{figure*}

\section{Method}
\label{sec:method}
We consider the problem of reconstructing world-space human motion from unconstrained video. An overview is given in Fig~\ref{fig:method}. The camera-space model encodes video features and generates human motion in camera coordinates. This motion is transformed to the world coordinates using camera poses, and becomes an initial noisy proposal for world-space motion. The world-space model encodes this noisy proposal and generates clean and coherent world-space human motion.   

\subsection{Motion representation}
We represent the human body as vertices on a 3D mesh $\mathbf{X} \in \mathbb{R}^{V\times 3}$. We denote the human body state at video time $t$ in coordinate system $a$ as ${}^a\!\mathbf{X}_t$. With this convention, the camera-space motion ($\mathbf{C}$) is the motion sequence where each frame's state is in its own instantaneous camera coordinates. The full camera-space motion sequence is 
\begin{equation}
\mathbf{C} = ({}^1\!\mathbf{X}_1, {}^2\!\mathbf{X}_2, ..., {}^T\!\mathbf{X}_T).
\end{equation}
For each video, we define a unique world coordinate system by setting the initial camera pose as the origin, and subsequently measuring all motion relative to it. The world-space human motion ($\mathbf{W}$) is a fixed coordinate system with respect to the camera of the first video frame,
\begin{equation}
\mathbf{W} = ({}^1\!\mathbf{X}_1, {}^1\!\mathbf{X}_2, ..., {}^1\!\mathbf{X}_T).
\end{equation}

The camera-space and world-space motion are related by camera poses at each timestep. Let $\mathbf{g}_t$ be the camera pose that maps the $t$-th camera coordinates to the world coordinates (camera 1). We can `lift' the camera-space motion $\mathbf{C}$ to world-space motion $\mathbf{W}$ using the camera motion
\begin{equation}
{}^1\!\hat{\mathbf{X}}_{t} = \mathbf{g}_t({}^t\!\mathbf{X}_t).
\label{eq:lift}
\end{equation}
where the hat operator denotes `lifted' variables. 

Our goal is to train diffusion models to generate $\mathbf{C}$ and $\mathbf{W}$. To ensure numerical stability, we decompose $\mathbf{X}$ as follows: in camera space, each state ${}^t\!\mathbf{X}_t$ is decomposed as the root-centered mesh ${}^t\!\mathbf{P}_t \in \mathbb{R}^{V\times3}$ and the root position ${}^t\!\mathbf{r}_t \in \mathbb{R}^{3}$ so that ${}^t\!\mathbf{X}_t = {}^t\!\mathbf{P}_t + {}^t\!\mathbf{r}_t$. Our camera-space model generates $({}^t\!\mathbf{P}_t, {}^t\!\mathbf{r}_t)$. In the world space, ${}^1\!\mathbf{r}_t$ is unbounded, growing larger with longer in-the-wild sequences, making it unstable for a diffusion model to learn directly. We instead model the velocity ${}^1\!\mathbf{v}_t = {}^1\!\mathbf{r}_t - {}^1\!\mathbf{r}_{t-1}$. The world-space mesh can be recovered using ${}^1\!\mathbf{X}_t = {}^1\!\mathbf{P}_t + \sum_{i=1}^{t}{}^1\!\mathbf{v}_i$. Our world-space diffusion model generates $({}^t\!\mathbf{P}_t, {}^t\!\mathbf{v}_t)$. For brevity, we still use $\mathbf{X}$ to denote the input and output of our diffusion models in the following sections.

\subsection{Camera-space diffusion}
\label{sec:cam}
We generate camera-space human motion $\mathbf{C}$ from video inputs. First, we extract two types of features: dense keypoints and image features. For each subject, we detect dense keypoints $\mathbf{L}_t \in \mathbb{R}^{V\times 3}$ (homogeneous coordinates) using a model similar to recent dense keypoint methods~\cite{camerahmr, cuevas2025mamma}. These dense keypoints have one-to-one correspondences with the 3D vertices on our sparse mesh. Instead of ingesting the keypoints' 2D coordinates directly, we propose to convert them to ray directions using the camera intrinsics $\mathbf{K}_t$. We then encode each ray using a positional embedding function $\gamma$~\cite{mildenhall2021nerf} and flatten the results, before processing them with an MLP. For invisible keypoints, we replace their embeddings with a learnable null embedding. This design implicitly encodes camera intrinsics, which is important for reconstruction accuracy~\cite{cliff, feng2025physhmr, multihmr}:
\begin{equation}
\begin{aligned}
\mathbf{E}_t &= \gamma(\mathbf{K}_{t}^{-1} \cdot \mathbf{L}_t) \\
\mathbf{f}^{\mathrm{kpt}}_t &= \mathrm{MLP}(\mathrm{vec}(\mathbf{E}_t)) \\
\mathbf{f}^{\mathrm{img}}_t &= \mathrm{Encoder}(\mathbf{I}_t) \\
\mathbf{f}_t &= \mathbf{f}^{\mathrm{kpt}}_t + \mathbf{f}^{\mathrm{img}}_t,
\end{aligned}
\end{equation}
where $\mathrm{vec}$ is the vectorization (flatten) operator. To extract image features, we input each frame $\mathbf{I}_t$ to an image encoder to get $\mathbf{f}^{\mathrm{img}}_t$. Finally, the two features are summed to get the final per-frame features $\mathbf{f}_t$. 

The camera-space motion diffusion, $\mathcal{D}_{\mathrm{cam}}$, takes this sequence of tokens $\mathbf{f}_{1:T}$ as conditioning and generates motion $\mathbf{C}$. It uses a standard DiT architecture~\cite{peebles2023scalable}, where conditions $\mathbf{f}_{1:T}$ are concatenated with noisy input $\mathbf{C}_\tau$ along the feature dimension, and processed with blocks of self-attention to predict the clean output $\mathbf{C}$: 
\begin{equation}
\begin{aligned}
\mathbf{C} &= \mathcal{D}_{\mathrm{cam}}(\mathbf{C}_\tau, \tau, \mathbf{f}_{1:T}),
\end{aligned}
\end{equation}
where $\tau\in\{1, ...,\mathcal{T}\}$ is the diffusion time step~\cite{ddpm}. 

We make a few improvements to the standard DiT to allow inference on longer sequences. First, we use relative positional embedding RoPE~\cite{su2024roformer} on the input sequence. Second, we apply windowed attention to the sequence via masking. Consequently, tokens at each time step only attend to a local temporal neighborhood. At inference, our model processes videos with thousands of frames without having to chunk.

\para{Height conditioning} As illustrated in Fig.~\ref{fig:height}, height of the human body remains a large source of ambiguity despite the increasing amount of synthetic training data. However, this information is readily available in many applications. Therefore, we train our model to incorporate this information when it's available. We show in experiments that learning to condition on height largely improves metric accuracy. To condition the model, we encode height (in meters) using an MLP and add it to the diffusion time step embedding. If height information is unavailable, we substitute the encoding with a learned null token~\cite{li2025genmo}.

\subsection{World-space diffusion}
\label{sec:world}
We generate globally consistent world-space motion from the initial noisy proposal (Eq.~\ref{eq:lift}) using our world-space motion diffusion model $\mathcal{D}_{\mathrm{world}}$. $\mathcal{D}_{\mathrm{world}}$ uses the same DiT architecture (see Sec.~\ref{sec:cam}), but takes the noisy lifted motion as conditioning. We encode ${}^1\!\hat{\mathbf{X}}_{t}$ with an MLP to get the condition $\mathbf{c}_{t}$ for each frame, 
\begin{equation}
\begin{aligned}
\mathbf{c}_{t}&= \mathrm{MLP}(\mathrm{vec}({}^1\!\hat{\mathbf{X}}_{t})), \\
\mathbf{W} &= \mathcal{D}_{\mathrm{world}}(\mathbf{W}_\tau, \tau, \mathbf{c}_{1:T}).
\end{aligned}
\end{equation}

\para{Masked modeling} To enable $\mathcal{D}_{\mathrm{world}}$ to generate plausible motion when the subject is unobservable, we incorporate temporal masking in the training. Specifically, $\mathbf{c}_t$ is randomly replaced with a learnable mask token during training, to simulate no visibility at time $t$. At inference, the visibility token can be toggled and is determined by the number of confident keypoints. 

\subsection{Guided sampling}
\label{sec: guidance}
While our dual models produce accurate motion, the world-space model's reliance on root velocity introduces two specific challenges that we address with test-time guided sampling. The first challenge is temporal drift. Because the world-space model outputs root velocities, integrating these velocities over time accumulates errors. This drift, while often minor, is observable as both 2D reprojection error and 3D misalignment with the scene, as visualized in Figure~\ref{fig:guidance} (right).

The second challenge arises during long-range occlusions, when a person is unobservable for an extended period (e.g., from frame $i$ to $j$).  While our world-space model can generate plausible connecting motion, it is not conditioned on the person’s absolute reappearance position. As a result, the integrated velocity $\sum {}^1\mathbf{v}_{i:j}$ may produce a trajectory that does not land at the correct destination, as visualized in Figure~\ref{fig:guidance} (left).

To address both issues, we introduce two test-time guidance terms that steer the sampling trajectory during inference. The first is a 2D reprojection guidance
\begin{equation}
\mathcal{L}_{\text{repro}} = \sum_{t=1}^T \| \mathbf{L}_t - \mathbf{K}_t{}\cdot \mathbf{g}_t^{-1}({}^1\!\mathbf{X}_t)\|,
\label{eq:guide2d}
\end{equation}
where $\mathbf{L}_t$ are the detected 2D dense landmarks and $\mathbf{g}_t^{-1}$ is the inverse camera pose. This loss encourages the world-space motion to align with the original video, to minimize temporal drift. 

The second guidance is a displacement guidance, applied specifically during long occlusions (e.g. $>$2 seconds),
\begin{equation}
\mathcal{L}_{\text{disp}} = \sum_{(i,j) \in \text{Occ}}\| ({}^1\hat{\mathbf{r}}_j - {}^1\hat{\mathbf{r}}_i) - \sum_{t=i}^j {}^1\mathbf{v}_t \|,
\label{eq:guide2d}
\end{equation}
where $\text{Occ}$ is the set of long-occlusion segments. This loss ensures that the total displacement from the integrated root velocities matches the displacement between the human's last visible position (${}^1\hat{\mathbf{r}}_i$) and first reappearance position (${}^1\hat{\mathbf{r}}_j$). These world-space positions are estimated by lifting the camera-space root position ${}^1\hat{\mathbf{r}}_t = g_t ({}^t\mathbf{r}_t)$.

We implement these guidance objectives as $x_0$-guidance during DDIM sampling~\cite{song2020ddim}. At each diffusion step, we compute the guidance gradient with respect to the predicted clean sample $\hat{x}_0$ and use it to update $\hat{x}_0$ before applying the standard DDIM update~\cite{bansal2023universal}. This approach is efficient as it avoids differentiating through the diffusion network~\cite{yuan2023physdiff}.

\subsection{Conversion to SMPLX}
\label{sec: conversion}
Prior work demonstrates regressing mesh vertices for individual frames~\cite{graphcmr, ho2025phd, meshform}, while our method generates vertices for the entire motion sequence. We convert the outputs to a parametric model for compactness and direct comparison with prior work.

We train an iterative MLP to convert our mesh to SMPLX parameters, avoiding the slow optimization-based conversion. This converter is inspired by learned optimization~\cite{learn2fit, learn2fit2, pymaf}. At each iteration, we compute the 3D errors between our sparse mesh vertices and the correspondences on the SMPLX mesh. The network takes the errors as input and predicts an update to the SMPLX parameters. Full details are in the supplementary material.

\begin{figure}[t!]
    \centering
    \includegraphics[width=0.98\linewidth]{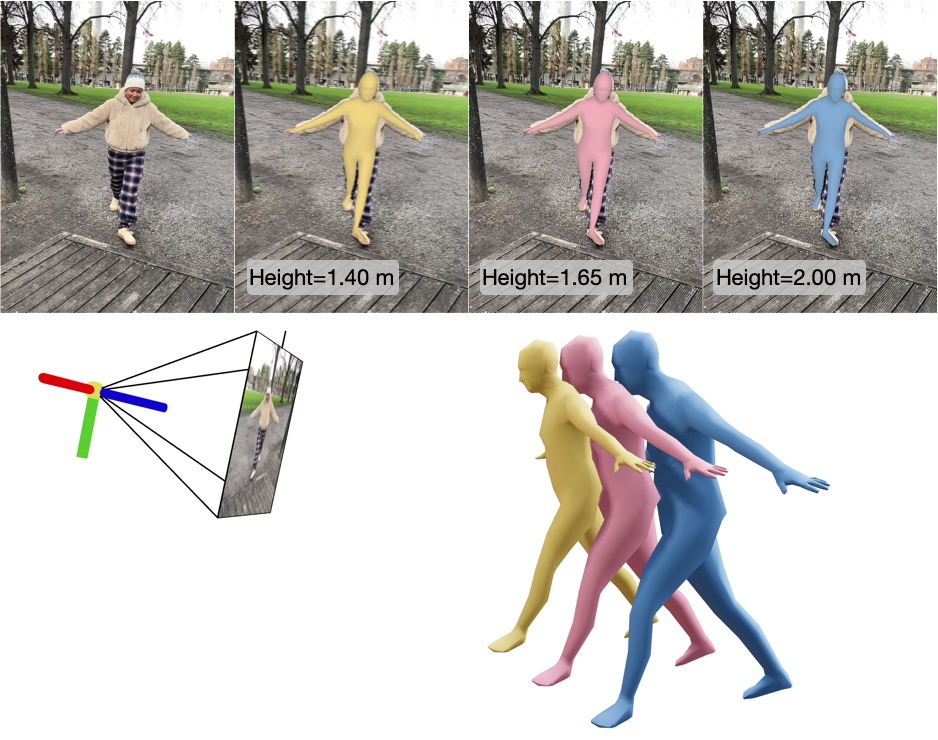}
    \vspace{-3mm}
    \caption{\textbf{Height conditioning}. Our camera-space model can generate predictions based on input body heights. As shown at the bottom row, height impacts distance from camera and thus plays an important role in world-space accuracy.}
    \label{fig:height}
    \vspace{-2mm}
\end{figure}

\subsection{Training}
\label{sec:train}

\para{Training losses} We train our camera-space and the world-space models separately. We train the camera-space model with a simple loss function
\begin{align*}
\mathcal{L}_{\text{Camera-space}}= \mathcal{L}_{\text{vertices}} + \mathcal{L}_{\text{position}} + \mathcal{L}_{\text{joints}},
\end{align*}
where $\mathcal{L}_{\text{vertices}}$, $\mathcal{L}_{\text{position}}$, and $\mathcal{L}_{\text{joints}}$ are L1 losses on the predicted mesh vertices ($P$), root position ($r$), and derived joint positions, respectively all in the camera space. 

After training, the model weights are frozen. We then use it in the training loop of the world-space model. The loss for the world-space model is defined as 
\begin{align*}
\mathcal{L}_{\text{World-space}}= \mathcal{L}_{\text{vertices}} + \mathcal{L}_{\text{velocity}}  + \mathcal{L}_{\text{contact}},
\end{align*}
which includes an L1 loss on the vertices and root velocity in the world space.  

The $\mathcal{L}_{\text{contact}}$ term is important for reducing foot-skating artifacts. Prior works predict contact labels for a post-hoc foot-locking step~\cite{shen2024gvhmr, wang2025prompthmr}. In contrast, we reinforce physically-plausible contact during training. We define this loss as the L1 error on the world-space foot vertices, applied only on frames where the foot is in contact with the ground:
\begin{align*}
\mathcal{L}_{\text{contact}}= \frac{1}{|S|}\sum_{t \in S} \|{}^1\!X_{t, \text{foot}} -  {}^1\!X^{*}_{t, \text{foot}}\|,
\label{eq:loss_contact}
\end{align*}
where ${}^1\!X_{t, \text{foot}}$ are the predicted foot vertices at $t$, ${}^1\!X^*_{t, \text{foot}}$ is the corresponding ground truth, and $S$ is the set of frames where the foot is in contact with the ground.

\para{Implementation details} 
We adopt the 595-vertex LOD6 sparse mesh from MHR~\cite{ferguson2025mhr}. This specific level of detail effectively captures human pose and shape, while the low vertex count makes training more efficient.

For our camera-space model, we use the frozen image encoder from PromptHMR~\cite{wang2025prompthmr}. The encoder from HMR2~\cite{hmr2} has a similar effect. We use the dense keypoint detection model from SAM 3D Body~\cite{yang2026sam}.

We trained our motion diffusion models separately with AdamW for one million steps, with a learning rate of $10^{-4}$ and an effective batch size of 256. During training, the sequence length $T$ is set to 120. Full implementation details are in the supplementary material.

\para{Datasets}
We trained our two motion diffusion models with common datasets. The camera-space model is trained with AMASS~\cite{amass}, BEDLAM~\cite{bedlam}, 3DPW~\cite{3dpw} and Goliath~\cite{goliath}. The world-space model is trained with AMASS~\cite{amass} and BEDLAM~\cite{bedlam}. We trained the dense keypoint detection model on BEDLAM. On AMASS, we simulate camera motion following WHAM~\cite{wham}.

\begin{table}[t!]
\centering
\setlength{\tabcolsep}{2pt}
\renewcommand{\arraystretch}{1.1}
\resizebox{0.49\textwidth}{!}
{\scriptsize{
\begin{tabular}{cl?ccc?ccc}
\cmidrule[0.75pt]{1-8}
& & \multicolumn{3}{c}{EMDB (24)} & \multicolumn{3}{c}{RICH (24)} \\
\cmidrule(lr){3-5} \cmidrule(lr){6-8}

& Models & \scriptsize{PA-MPJPE} & \scriptsize{MPJPE} & \scriptsize{PVE} & \scriptsize{PA-MPJPE} & \scriptsize{MPJPE} & \scriptsize{PVE}  \\
\cmidrule{1-8}

\multirow{5}{1em}{\rotatebox[origin=c]{90}{\tiny{per-frame}}} 

& HMR2.0~\cite{hmr2} & 61.5 & 97.8 & 120.0 & 60.7 & 98.3 & 120.8 \\

& ReFit~\cite{tokenhmr}  & 58.6 & 88.0 & 104.5 & 47.9 & 80.7 & 92.9 \\

& CameraHMR~\cite{camerahmr}  & 43.3 & 70.2 & 81.7 & \textbf{34.0} & 55.7 & 64.4 \\

&NLF~\cite{nlf} & 41.2 & 69.6 & 82.4 & -- & -- & -- \\

&PromptHMR~\cite{wang2025prompthmr} & \textbf{41.0} & 71.7 & 84.5 & 37.3 & 56.6 & 65.5 \\

\cmidrule{1-8}

\multirow{6}{1em}{\rotatebox[origin=c]{90}{\tiny{temporal}}} 

& WHAM~\cite{wham} & 52.0 & 81.6 & 96.9 & 44.3 & 80.0 & 91.2 \\
& TRAM~\cite{wang2024tram} & 45.7 & 74.4 & 86.6 & - & - & - \\
& GVHMR~\cite{shen2024gvhmr} & 44.5 & 74.2 & 85.9 & 39.5 & 66.0 & 74.4 \\
& GENMO~\cite{li2025genmo} & 42.5 & 73.0 & 84.8 & 39.1 & 66.8 & 75.4 \\
\cmidrule{2-8}

& \textbf{Ours} & 41.8 &  \underline{67.1} &  \underline{78.2} & \underline{34.8} &  \underline{51.4} &  \underline{58.9} \\
& \textbf{Ours (w/ height)} & \underline{41.7} & \textbf{59.5} & \textbf{70.4} & 35.0 & \textbf{48.0} & \textbf{55.2} \\
\cmidrule[0.75pt]{1-8}
\end{tabular}
}}
\vspace{-2mm}
\caption{\textbf{Camera-space reconstruction} on EMDB~\cite{emdb} and RICH~\cite{huang2022capturing}, with the number of joints in parenthesis. Our method does not use test-time flip augmentation. All metrics are in $mm$. }
\vspace{-3mm}
\label{tab:cam_space}
\end{table}

\section{Experiments}
\label{sec:exp}
We evaluate DuoMo on camera-space reconstruction, world-space reconstruction, and motion quality. To compare with prior methods, we use our converter to transform our sparse meshes to SMPLX meshes and joints to compute accuracy. Our method uses ground truth camera intrinsics for all evaluations. 

\subsection{Camera-space reconstruction}
We first evaluate our camera-space model's accuracy on the EMDB~\cite{emdb} and RICH~\cite{huang2022capturing} datasets. As shown in Table~\ref{tab:cam_space}, our method achieves new state-of-the-art performance.

We find this accuracy is impacted by the inherent scale ambiguity of monocular reconstruction. This is visualized in Figure~\ref{fig:height}, which shows that multiple 3D reconstructions can align perfectly with the 2D image evidence while having different body heights and distances from the camera.

Our height-conditioning mechanism resolves this ambiguity with additional information. Conditioning on ground-truth subject height during inference yields a 10\% improvement on both MPJPE and PVE. This result demonstrates that resolving body scale is a crucial step for accurately ``lifting" the motion into a consistent world space.

\begin{table}[ht]
\centering
\setlength{\tabcolsep}{3pt}
\renewcommand{\arraystretch}{1.1}
\resizebox{0.47\textwidth}{!}
{\scriptsize{
\begin{tabular}{cl?cc?cccc}
\cmidrule[0.75pt]{1-8}
& & \multicolumn{2}{c}{Visible segment} & \multicolumn{4}{c}{Full segment} \\
\cmidrule(lr){3-4} \cmidrule(lr){5-8} 

& Models & \scriptsize{W-MPJPE} & \scriptsize{RTE} & \scriptsize{W-MPJPE} & \scriptsize{RTE} & \scriptsize{W-MPJPE-Occ} & \scriptsize{RTE-Occ} \\
\cmidrule{1-8}

& PHMR~\cite{wang2025prompthmr} & 271.1 & 2.3 & 357.2 & 3.3 & 885.0 & 12.0 \\
& GVHMR~\cite{shen2024gvhmr} & 189.1 & 3.9 & 209.4 & 3.9 & 384.1 & 6.1 \\
\cmidrule{2-8}
& Cam-model + Lifting & 95.3 & \textbf{0.7} & 174.2 & 1.9 & 688.1 & 10.4 \\
& DuoMo & 99.7 & 1.5 & 109.3 & 1.6 & 193.1 & 2.6 \\
& DuoMo w/ guidance & \textbf{90.4} & 1.0 & \textbf{101.3} & \textbf{1.1} & \textbf{175.4} & \textbf{1.7} \\
\cmidrule[0.75pt]{1-8}
\end{tabular}
}}
\vspace{-2mm}
\caption{\textbf{Robust reconstruction} on the Egobody~\cite{ego_body}, evaluating both visible and invisible segments (e.g. person out of frame). All methods use ground truth camera poses in this evaluation.}
\vspace{-1mm}
\label{tab:egobody}
\end{table}

\begin{table}[t!]
\centering
\setlength{\tabcolsep}{3pt}
\renewcommand{\arraystretch}{1.1}
\resizebox{0.47\textwidth}{!}
{\scriptsize{
\begin{tabular}{c?ccccc}
\cmidrule[0.75pt]{1-6}
Methods & $\text{WA-MPJPE}$ & $\text{W-MPJPE}$ & RTE & Jitter & Foot Skating \\
\cmidrule(lr){1-6} 
World-space model (one stage) & 153.5 & 445.1 & 6.7 & \underline{9.1} &  \underline{4.8} \\
Cam-space model + Lifting &  \underline{67.0} &  \underline{180.2} &  \underline{1.3} & 32.6 & 9.2 \\
DuoMo  & \textbf{66.0} & \textbf{167.1} & \textbf{1.1} & \textbf{8.7} & \textbf{3.7} \\
\cmidrule[0.75pt]{1-6}
\end{tabular}
}}
\vspace{-2mm}
\caption{\textbf{Ablation for DuoMo design} on EMDB~\cite{emdb}. DuoMo has higher accuracy and motion quality than using only one model.}
\label{tab:ablation_duo}
\vspace{-4mm}
\end{table}

\begin{table*}[t!]
\centering
\footnotesize
\setlength{\tabcolsep}{3pt}
\begin{tabularx}{0.9\textwidth}{c l X c c c c c X c c c c c}
\cmidrule[0.75pt]{1-14}
% Header row
& & & \multicolumn{5}{c}{EMDB (24)} & & \multicolumn{5}{c}{RICH (24)} \\
\cmidrule(lr){4-8} \cmidrule(lr){10-14}
& Models & & \scriptsize{WA-MPJPE} & \scriptsize{W-MPJPE} & \scriptsize{RTE} & \scriptsize{Jitter} & \scriptsize{Foot-sliding} & & \scriptsize{WA-MPJPE} & \scriptsize{W-MPJPE} & \scriptsize{RTE} & \scriptsize{Jitter} & \scriptsize{Foot-sliding} \\
\cmidrule{1-14}
& TRACE~\cite{trace} & & 529.0 & 1702.3 & 17.7 & 2987.6 & 370.7 & & 238.1 & 925.4 & 610.4 & 1578.6  & 230.7 \\
& GLAMR~\cite{glamr} & & 280.8 & 726.6 & 11.4 & 46.3 & 20.7 & & 129.4 & 236.2 & 3.8 & 49.7 & 18.1 \\
& SLAHMR~\cite{slahmr} & & 326.9 & 776.1 & 10.2 & 31.3 & 14.5 & & 98.1 & 186.4 & 28.9 & 34.3 & 5.1 \\
& WHAM~\cite{wham} & & 135.6 & 354.8 & 6.0 & 22.5 & 4.4 & & 109.9 & 184.6 & 4.1 & 19.7 & 3.3 \\
& GVHMR~\cite{shen2024gvhmr} & & 111.0 & 276.5 & 2.0 & 16.7 & \textbf{3.5} & & 78.8 & 126.3 & 2.4 & 12.8 & \textbf{3.0} \\
& TRAM~\cite{wang2024tram} & & 76.4 & 222.4 & 1.4 & 18.1 & 11.0 & & -- & -- & -- & -- & -- \\
& GENMO~\cite{li2025genmo} & & 74.3 & 202.1 & 1.2 & 17.8 & 8.8 & & 75.3 & 118.6 & 1.9 & 15.0 & 6.7 \\
\cmidrule{2-14}
& \textbf{Ours} & & 67.6 & 171.6 & 1.2 & 8.7 & 3.7 & & 53.6 & 80.8 & 1.3 & 5.3 & 3.1 \\
& \textbf{Ours (w/ height)} & & \textbf{66.0} & \textbf{167.1} & \textbf{1.1} & \textbf{8.7} & 3.7 & & \textbf{53.5} & \textbf{80.4} & \textbf{1.3} & \textbf{5.2} & 3.1 \\
\cmidrule[0.75pt]{1-14}
\end{tabularx}
\vspace{-2mm}
\caption{\textbf{World-space reconstruction} on the EMDB~\cite{emdb} and RICH~\cite{huang2022capturing} datasets. We do not use test-time flip augmentation. Our method uses the same estimated camera motion as TRAM~\cite{wang2024tram} and GENMO~\cite{li2025genmo}, providing a meaningful comparison.}
\vspace{-3mm}
\label{tab:world_space}
\end{table*}

\begin{figure*}[t!]
    \centering
    \includegraphics[width=0.88\linewidth]{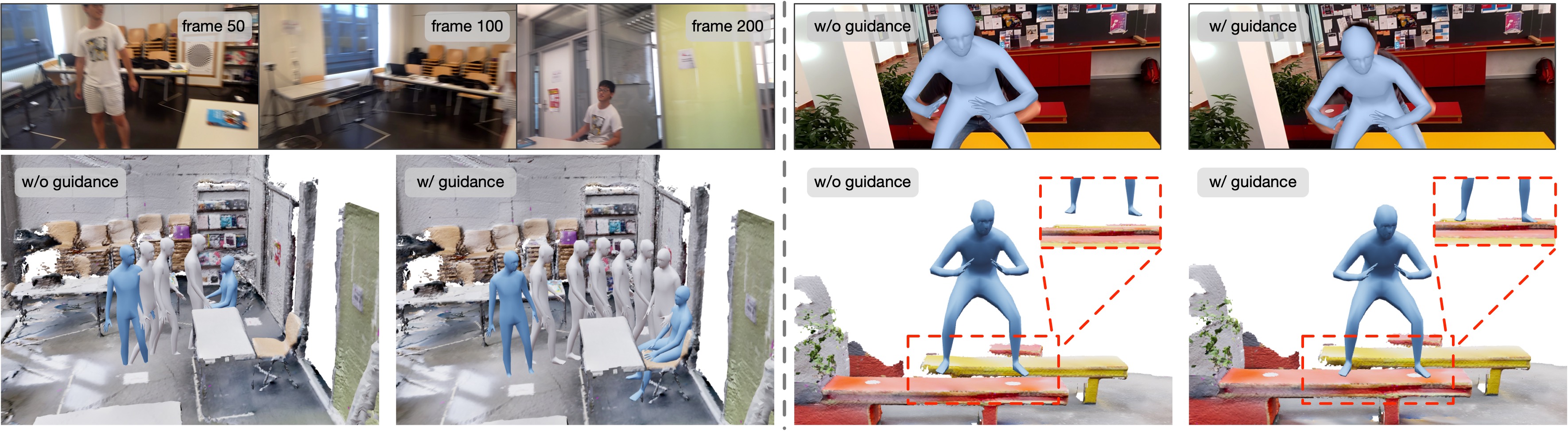}
    \vspace{-2mm}
    \caption{\textbf{Guided sampling} on Egobody~\cite{egobody}. Our proposed guidances correct for drifting and improve world-space trajectory accuracy.}
    \label{fig:guidance}
    \vspace{-2mm}
\end{figure*}

\subsection{World-space reconstruction}
We evaluate our world-space model's accuracy using the EMDB~\cite{emdb} and RICH~\cite{huang2022capturing} datasets. For a fair comparison on EMDB, we use the camera motion provided by TRAM~\cite{wang2024tram}, as most methods are sensitive to camera pose accuracy. On RICH, we use the static camera poses. 

We evaluate world-space motion using: WA-MPJPE (joint accuracy on Procrustes-aligned 100-frame segments); W-MPJPE (joint accuracy on 100-frame segments aligned only at the first two frames); and RTE (root trajectory position error in $\%$). Full details in the supplementary.

As shown in Table~\ref{tab:world_space}, DuoMo demonstrates significant improvements in both 3D accuracy and motion quality. Our W-MPJPE is 16\% lower on EMDB and 30\% lower on RICH than the second best method. At the same time, we have high motion quality, with low foot-sliding score comparable to GVHMR which requires post-hoc foot-locking.

\begin{figure*}[t!]
    \centering
     \includegraphics[width=0.90\linewidth]{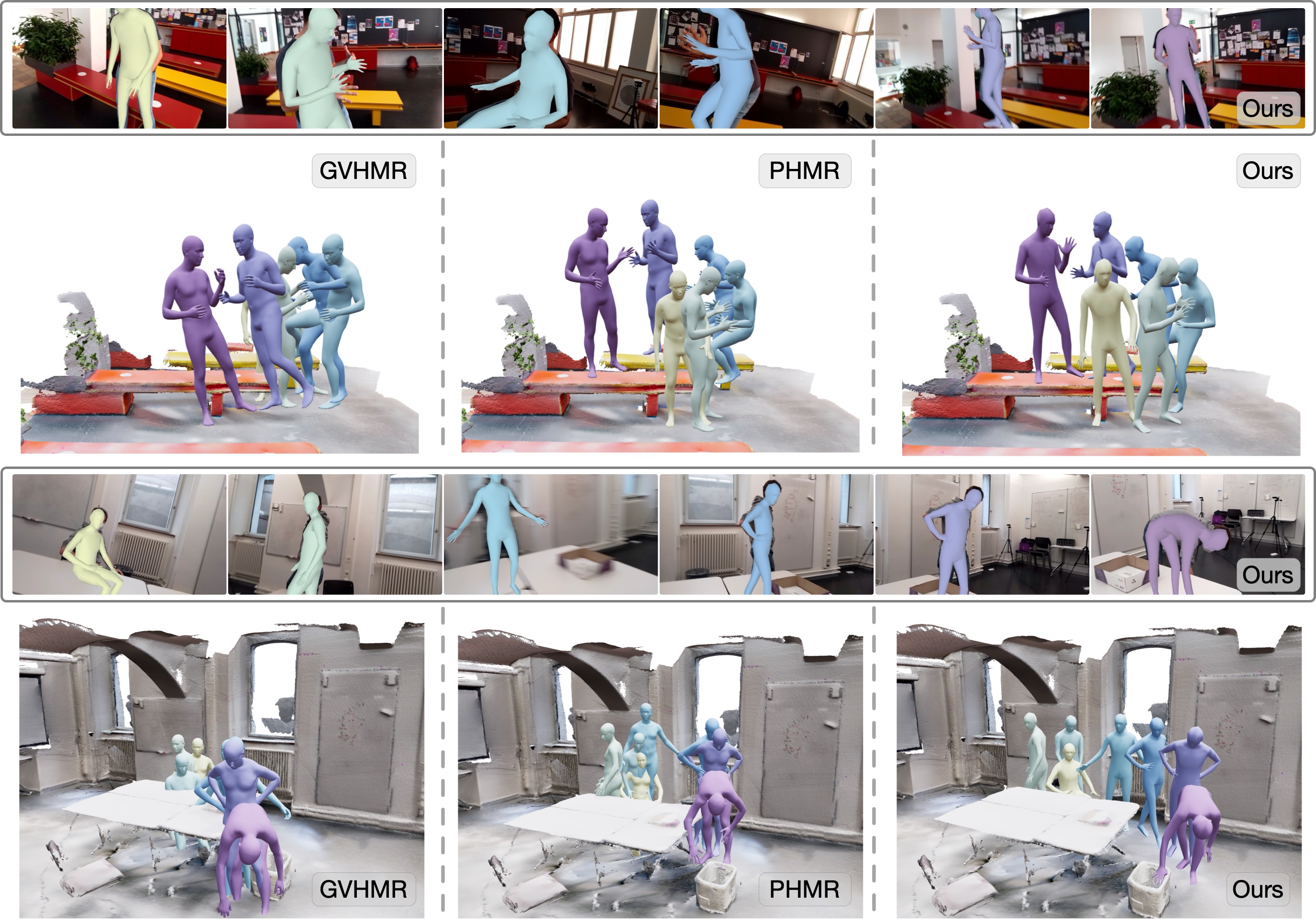}
    \vspace{-2mm}
    \caption{\textbf{Qualitative comparison} on Egobody~\cite{ego_body}. All methods use the ground truth camera poses. We observe that results from GVHMR~\cite{shen2024gvhmr} are smooth but drift under shaky camera motion. PromptHMR~\cite{wang2025prompthmr} has better position accuracy but is not robust to occlusion and depth ambiguity. Our results show both accuracy and robustness.}
    \label{fig:qualitative}
    \vspace{-2mm}
\end{figure*}

We show a qualitative comparison in Figure~\ref{fig:qualitative}. Compared to others~\cite{shen2024gvhmr, wang2025prompthmr}, our results have higher scene consistency and less drifting. We attribute this performance gain to our dual model design, which we analyze in our ablation studies~\ref{sec:ablation}.

\subsection{Generative reconstruction}
We evaluate the robustness and generative capabilities of DuoMo on the challenging egocentric subset of the Egobody~\cite{ego_body} dataset. These sequences are captured by a moving HoloLens featuring moving subjects in indoor environments. The videos are shaky, with subjects frequently truncated and out of frame. For this experiment, we use the tracked camera motion from HoloLens.

We establish two evaluation tracks. \textbf{Visible reconstruction} evaluates accuracy only on the frames where the subject is visible. \textbf{Full-sequence reconstruction} evaluates accuracy over the entire sequence, including when the subject is out of frame. For this track, we report metrics on the full sequence and also on the occluded segments specifically (W-MPJPE-Occ and RTE-Occ).

As shown in Table~\ref{tab:egobody}, DuoMo demonstrates high accuracy and robustness. A key comparison is against our camera-space model+lifting baseline. The baseline is unable to generate plausible world-space motion during occlusions, showing hight W-MPJPE-Occ error. In contrast, DuoMo's world-space modeling provides a generative prior to complete the motion, as evidenced by the gains on the full-sequence metrics.

\begin{table}[t!]
\centering
\setlength{\tabcolsep}{3pt}
\renewcommand{\arraystretch}{1.1}
\resizebox{0.47\textwidth}{!}
{\scriptsize{
\begin{tabular}{c?ccccc}
\cmidrule[0.75pt]{1-6}
DuoMo w/ & $\text{WA-MPJPE}$ & $\text{W-MPJPE}$ & RTE & Jitter & Foot Skating \\
\cmidrule(lr){1-6} 
%\cmidrule(lr){1-6} 
World-Model-SMPL & 70.1 & 182.5 &  1.3 &  \textbf{7.4} & 4.8 \\
World-Model-Mesh  & \textbf{65.7} & \textbf{164.8} & \textbf{1.1} & 8.5 & \textbf{3.8} \\
\cmidrule[0.75pt]{1-6}
\end{tabular}
}}
\vspace{-2mm}
\caption{\textbf{Ablation for motion representation} on EMDB~\cite{emdb}. Directly generating sparse mesh is very competitive in terms of motion reconstruction accuracy.}
\label{tab:smpl}
\vspace{-5mm}
\end{table}

Furthermore, these results underscore the importance of guided sampling. While the generative model can produce plausible motion, it is not sufficient for accurate reconstruction under occlusion. The generative process mush also consider the human's reappearance. Our 2D reprojection and displacement guidance terms enforce this, improving RTE and RTE-Occ. A qualitative comparison in Figure~\ref{fig:guidance} visualizes this effect.

\subsection{Ablation studies}
\label{sec:ablation}
\para{Dual priors} We perform ablation studies to validate the core technical choices. We first analyze the dual models in Table~\ref{tab:ablation_duo}. The direct lifting approach (Cam-model+Lifting) achieves good accuracy but its motion quality is poor with high jitter and foot skating. Conversely, we created a one-stage world-space model only baseline. This model learns a stronger world-space motion prior and produces plausible motion but has subpar accuracy. This highlights the difficulty of solving 2D-to-3D lifting and global consistency simultaneously. Prior works that attempt this often rely on stronger inductive biases~\cite{wham, shen2024gvhmr}. Our DuoMo factorization achieves the best of both worlds.

\para{Representation}
We freeze our camera-space model and train a new world-space model that outputs SMPLX parameters. We adapt the loss setup from prior works~\cite{hmr2}, adding our $\mathcal{L}_{\text{contact}}$ term for a fair comparison. As shown in Table~\ref{tab:smpl}, our sparse mesh approach is highly competitive. This is a promising outcome, as it suggests our architecture can be extended to learn motion models for other categories directly from surface supervision without parametric models.

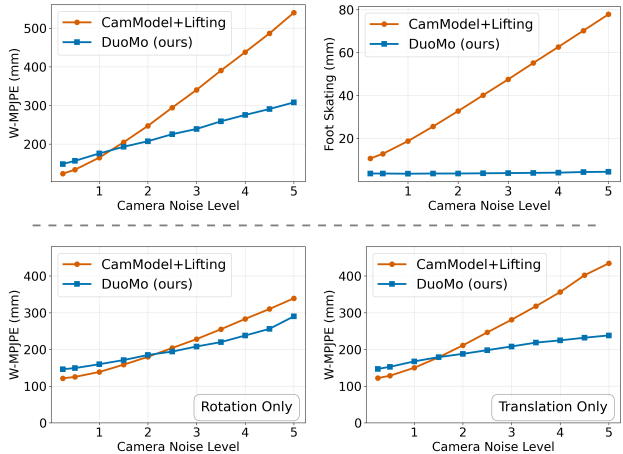
\begin{figure}[t]
    \centering
    \vspace{-1mm}
    % --- Top row ---
    \begin{subfigure}[b]{0.235\textwidth}
        \centering
        \includegraphics[width=\textwidth]{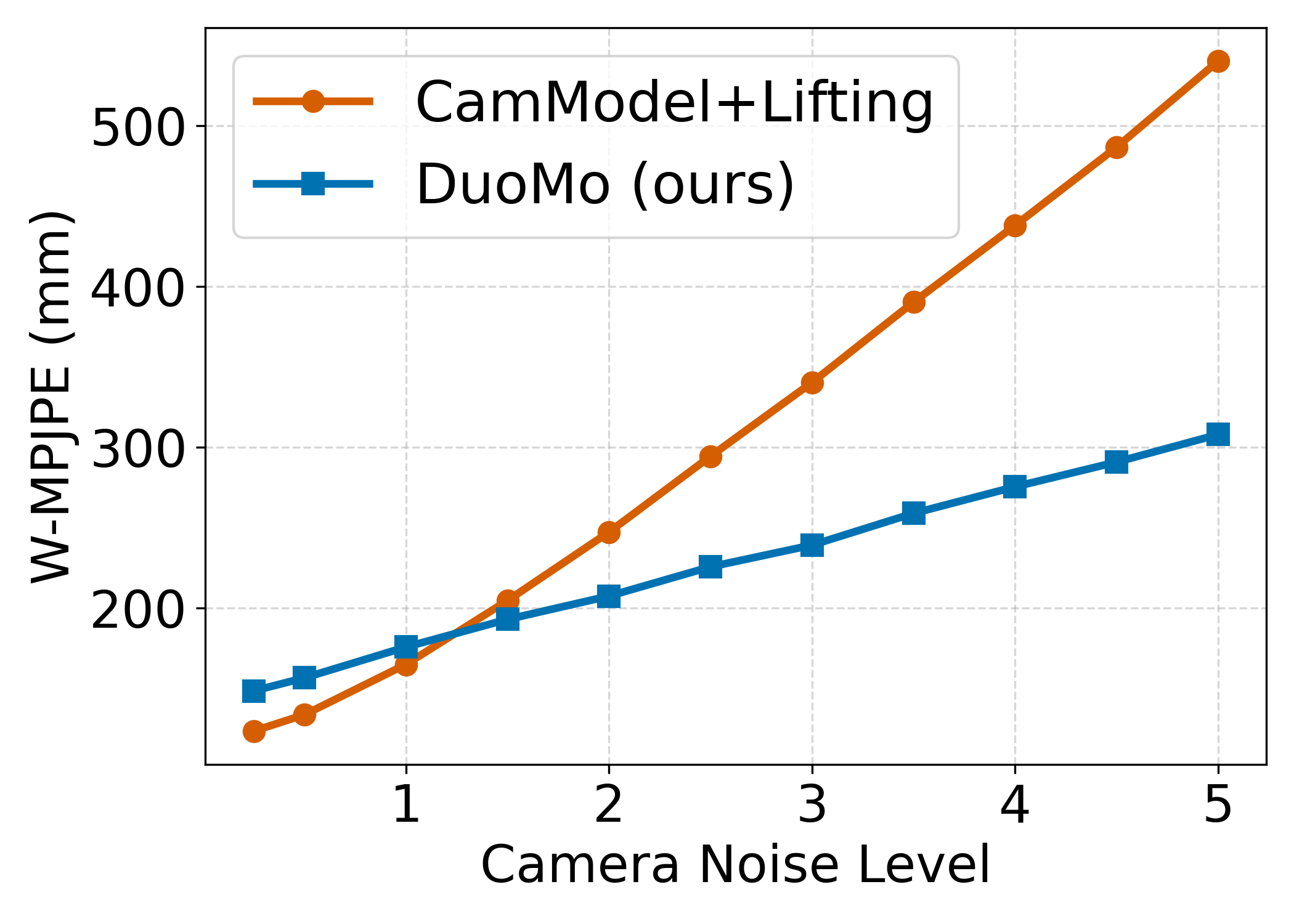}
        \label{fig:wmpjpe}
    \end{subfigure}
    \begin{subfigure}[b]{0.235\textwidth}
        \centering
        \includegraphics[width=\textwidth]{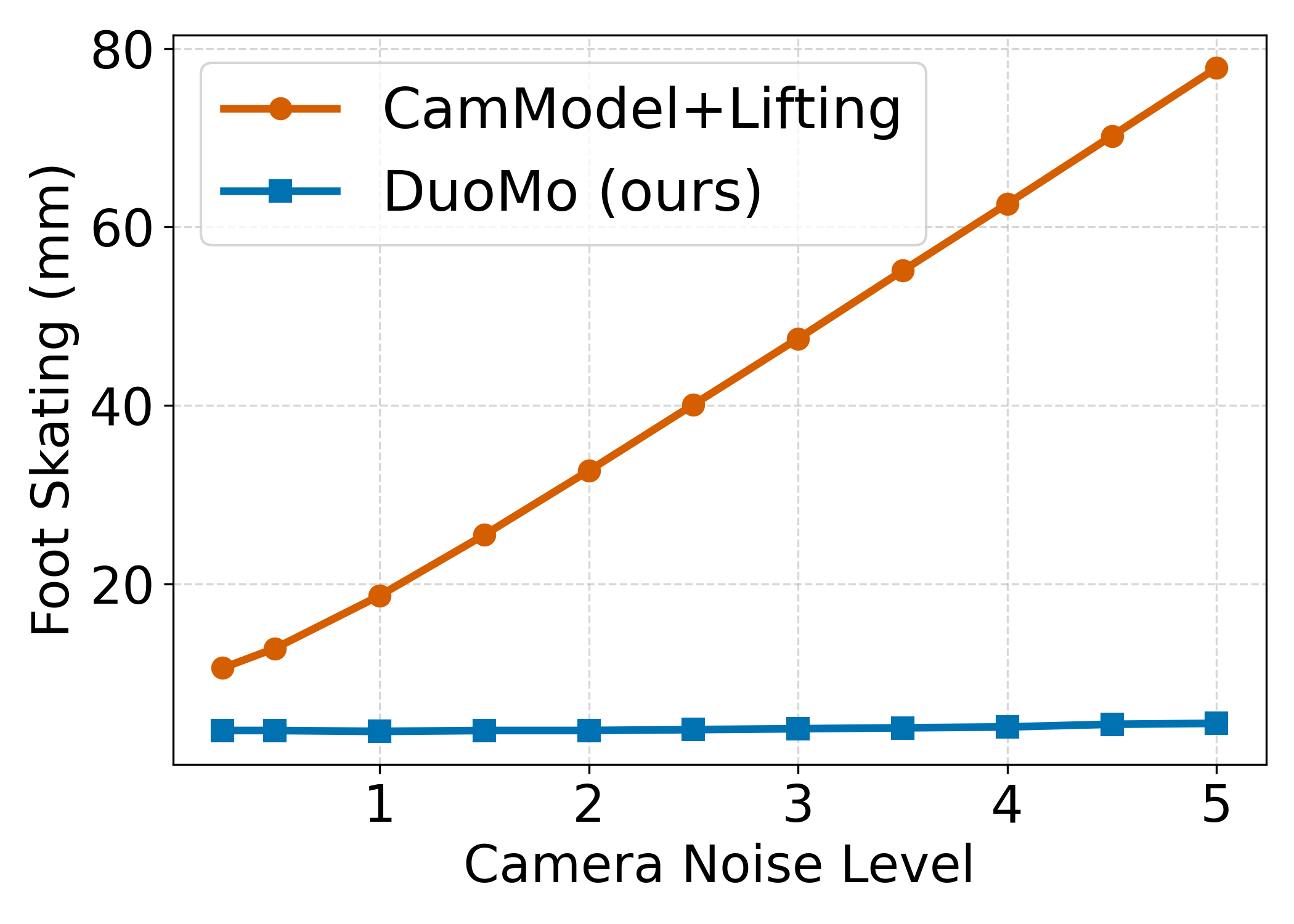}
        \label{fig:foot_skating}
    \end{subfigure}
    % --- Horizontal line between rows ---
    \par\vspace{-11mm}
\begin{center}
\begin{tikzpicture}
    \draw[dashed, gray, line width=0.7pt] (0,0) -- (7.5,0); % Adjust 5.5 for desired width
\end{tikzpicture}
\end{center}
    \par\vspace{-2mm}
    % --- Bottom row ---
       \begin{subfigure}[b]{0.235\textwidth}
        \centering
        \includegraphics[width=\textwidth]{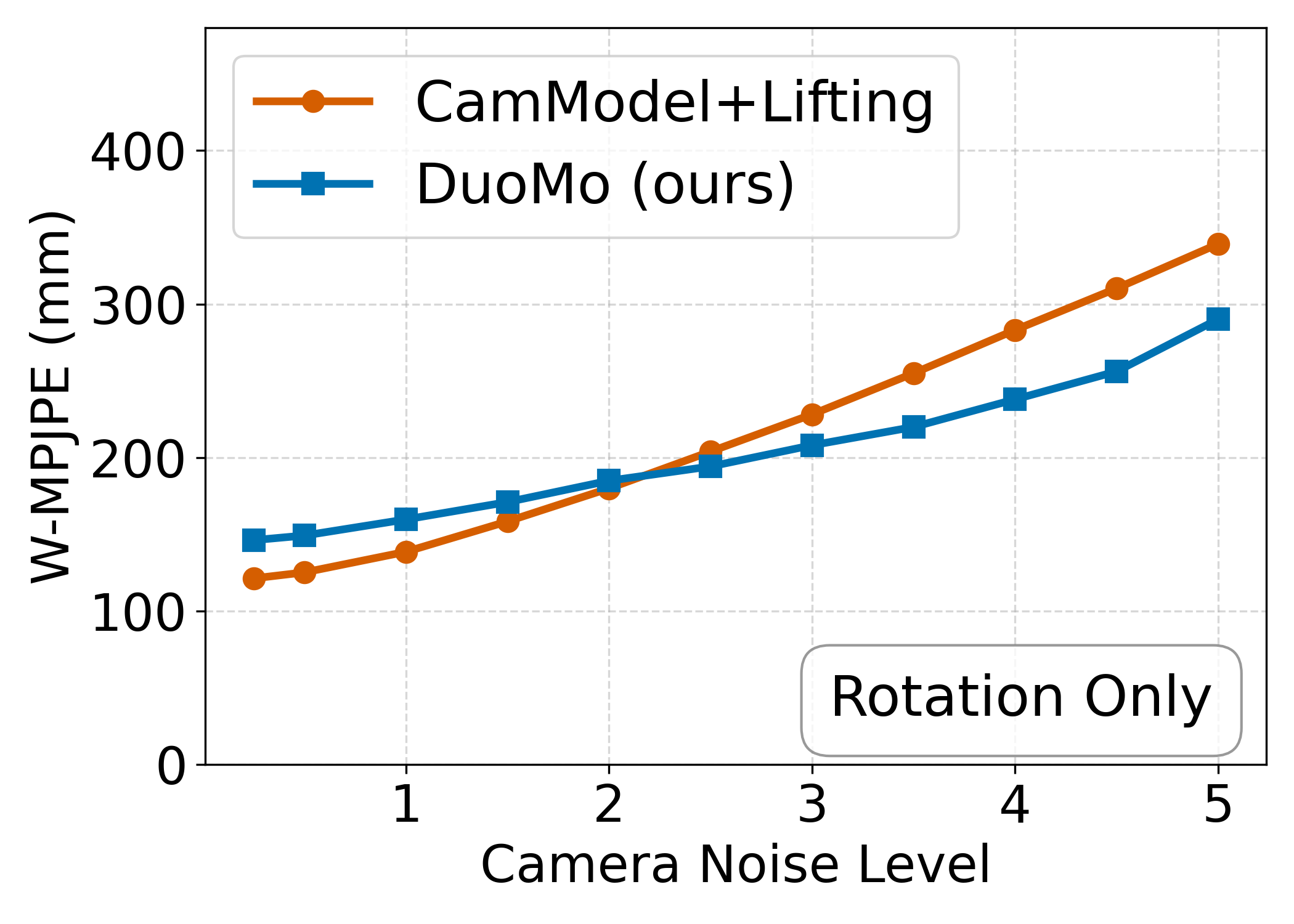}
        \label{fig:wmpjpe}
    \end{subfigure}
    \begin{subfigure}[b]{0.235\textwidth}
        \centering
        \includegraphics[width=\textwidth]{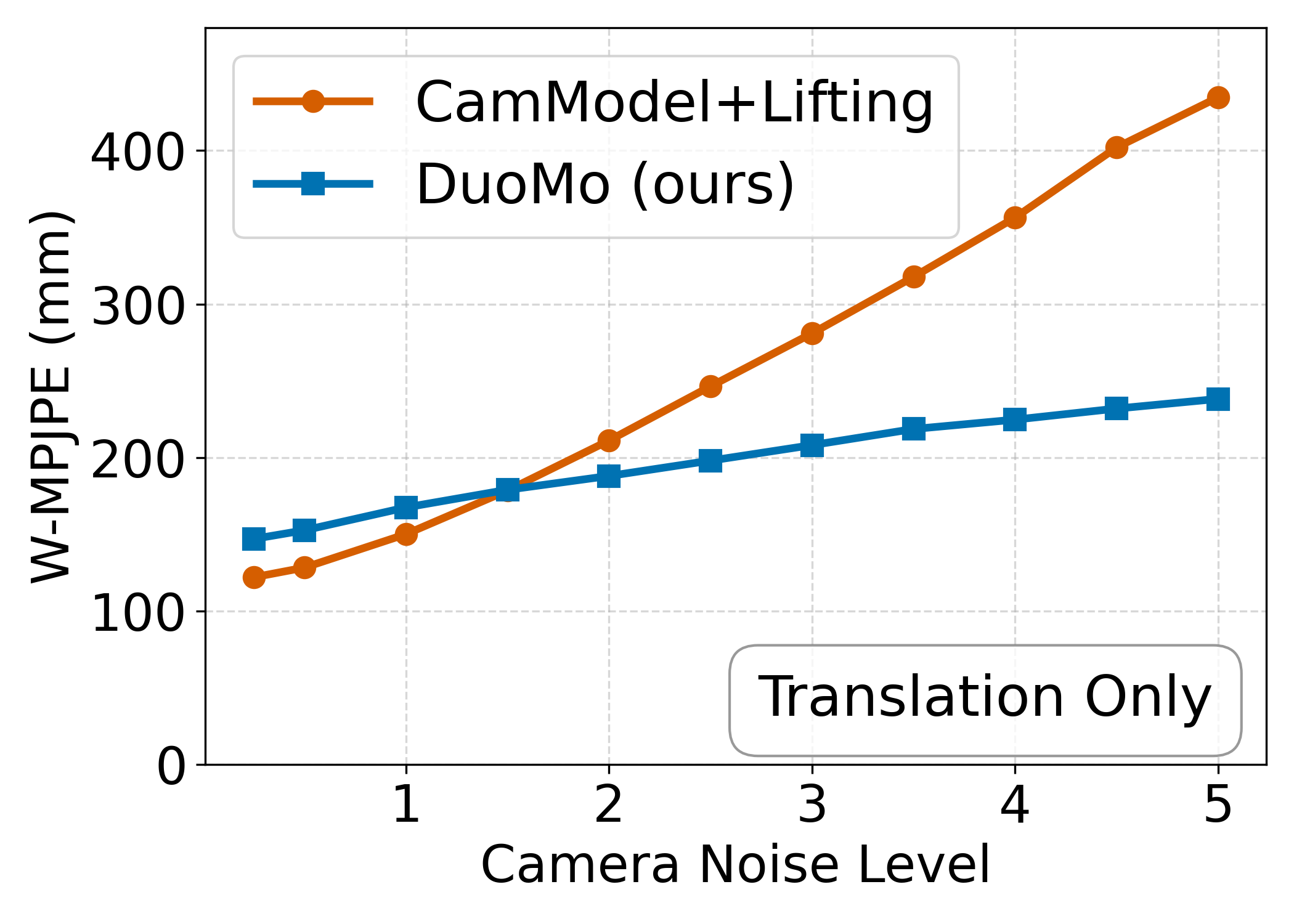}
        \label{fig:foot_skating}
    \end{subfigure}
    \vspace{-10mm}
    \caption{\textbf{Impact of camera error}. Comparison of methods under different camera motion noise levels.}
    \label{fig:ablation_cam}
    \vspace{-5mm}
\end{figure}

\para{Camera estimation error} We analyze the critical role of camera pose accuracy in world-space human reconstruction. First, we establish a realistic noise profile by comparing TRAM's camera estimation against the iPhone-tracked ``ground truth" camera poses on EMDB~\cite{emdb}. We then model this error as accumulating, per-frame Gaussian noise on the Lie algebra (for rotation) and Euclidean coordinates (for translation) to simulate SLAM drift. We apply this noise at increasing intensities to the ground truth camera. We sample 10 camera trajectories per-level per-sequence, and average our human reconstruction accuracy. 

Figure~\ref{fig:ablation_cam} (top) shows that the Cam-model+Lifting baseline, while achieving a low W-MPJPE at near-zero noise, is very brittle; its accuracy degrades rapidly, and its foot-skating (FS) score explodes as noise increases. In contrast, DuoMo is more robust. Its W-MPJPE degrades gracefully, and its foot-skating remains minimal across all noise levels.

This demonstrates that the world-space model acts as a generative regularizer: it sacrifices a small amount of per-frame fidelity at low noise to gain robustness and maintain physical plausibility. The bottom plot further isolates this effect, by only applying rotation noise or translation noise to the camera trajectory. It shows that while the Cam-model+Lifting baseline's error degrades rapidly from translation noise, DuoMo's accuracy is more stable.

\para{Speed} On an H200, a 20s video (30FPS) takes 2s, 3s, 30s, and 1.5s for extracting keypoints, dense keypoints, image features, and diffusion respectively. We discuss limitations in the supplementary.

\section{Conclusion}
\label{sec:conclusion}
We presented DuoMo, a two-stage diffusion method for world-space human motion reconstruction. Our approach achieves robustness and generality by decoupling the problem into a camera-space estimation stage and a world-space refinement stage that operates in per-video coordinate systems. Furthermore, our architecture generates the motion of mesh vertices directly. Experiments validate that DuoMo achieves state-of-the-art performance.

% \newpage
{
    \small
    \bibliographystyle{ieeenat_fullname}
    \bibliography{main}
}

\clearpage
\maketitlesupplementary
\appendix

\section{Sparse mesh representation}
\label{sec:sparse_mesh}

\subsection{Topology}
We use a sparse mesh $\mathbf{X} \in \mathbb{R}^{V\times 3}$ to represent the human body and its motion. The sparse mesh has $V=595$ vertices~\cite{ferguson2025mhr}, and is designed to balance between representation quality and efficiency. Figure~\ref{fig:topology} shows the sparse mesh topology. Specifically, the vertices are positioned to ensure that body pose, shape, and hand gestures are accurately represented despite the sparsity. Consistent with prior methods~\cite{smpl, smplx}, skeletal joints are regressed from the posed mesh using linear combinations of local vertices.

\subsection{SMPLX to sparse mesh}
\label{sec:smpl_to_sparse}
To leverage existing datasets for training our motion models, we create regression matrices to convert SMPL or SMPLX meshes to the sparse mesh, using a strategy similar to learning the joint regressor for SMPL~\cite{smpl}. Each vertex on the sparse mesh can be computed as a weighted average of a sparse set of vertices from the SMPL/SMPLX models. 

\begin{figure}[t]
    \centering
    \vspace{-1mm}
    % --- Top row ---
    \begin{subfigure}[b]{0.235\textwidth}
        \centering
        \includegraphics[width=\textwidth]{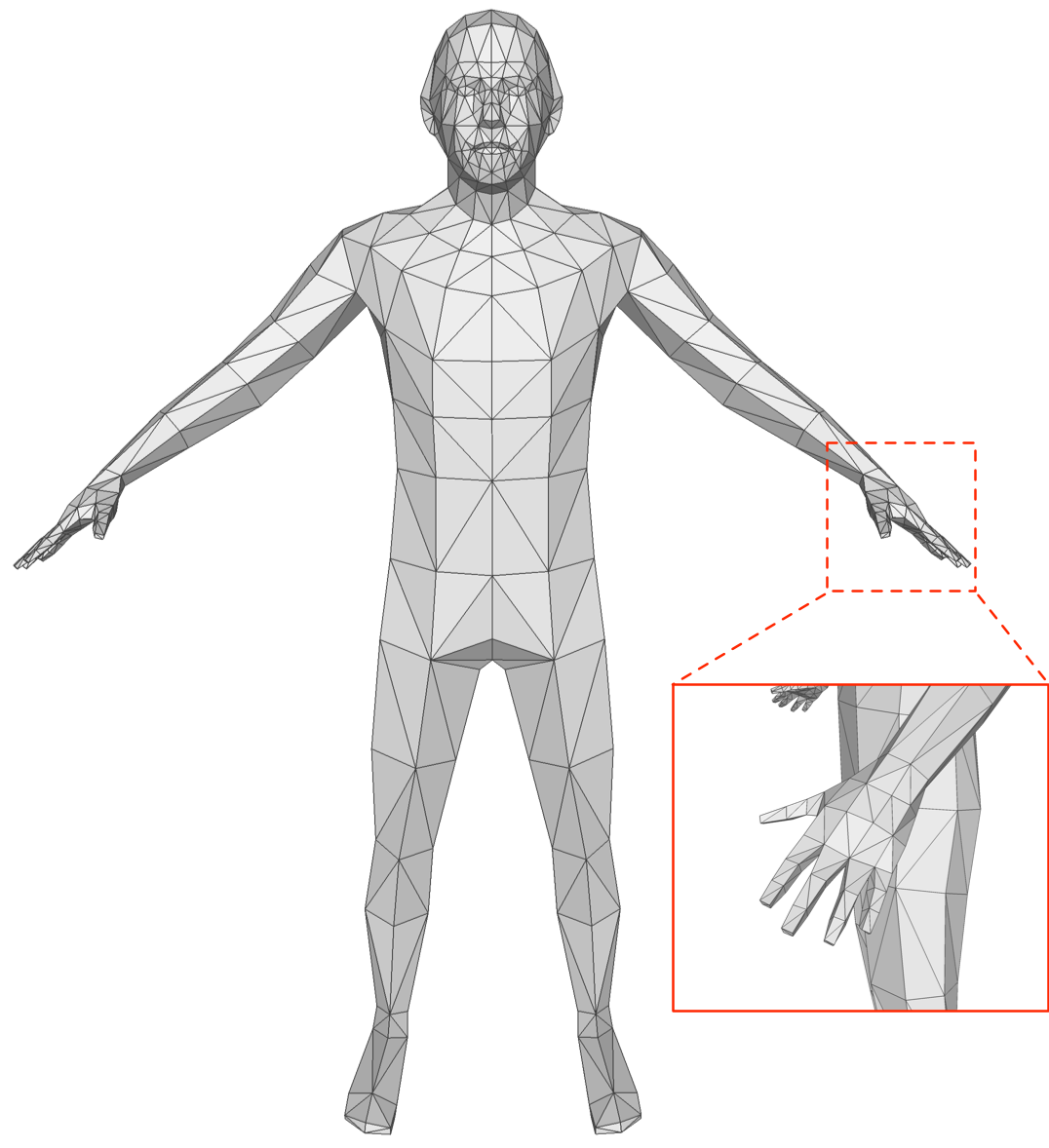}
        \caption{Topology}
        \label{fig:topology}
    \end{subfigure}
    \begin{subfigure}[b]{0.235\textwidth}
        \centering
        \includegraphics[width=\textwidth]{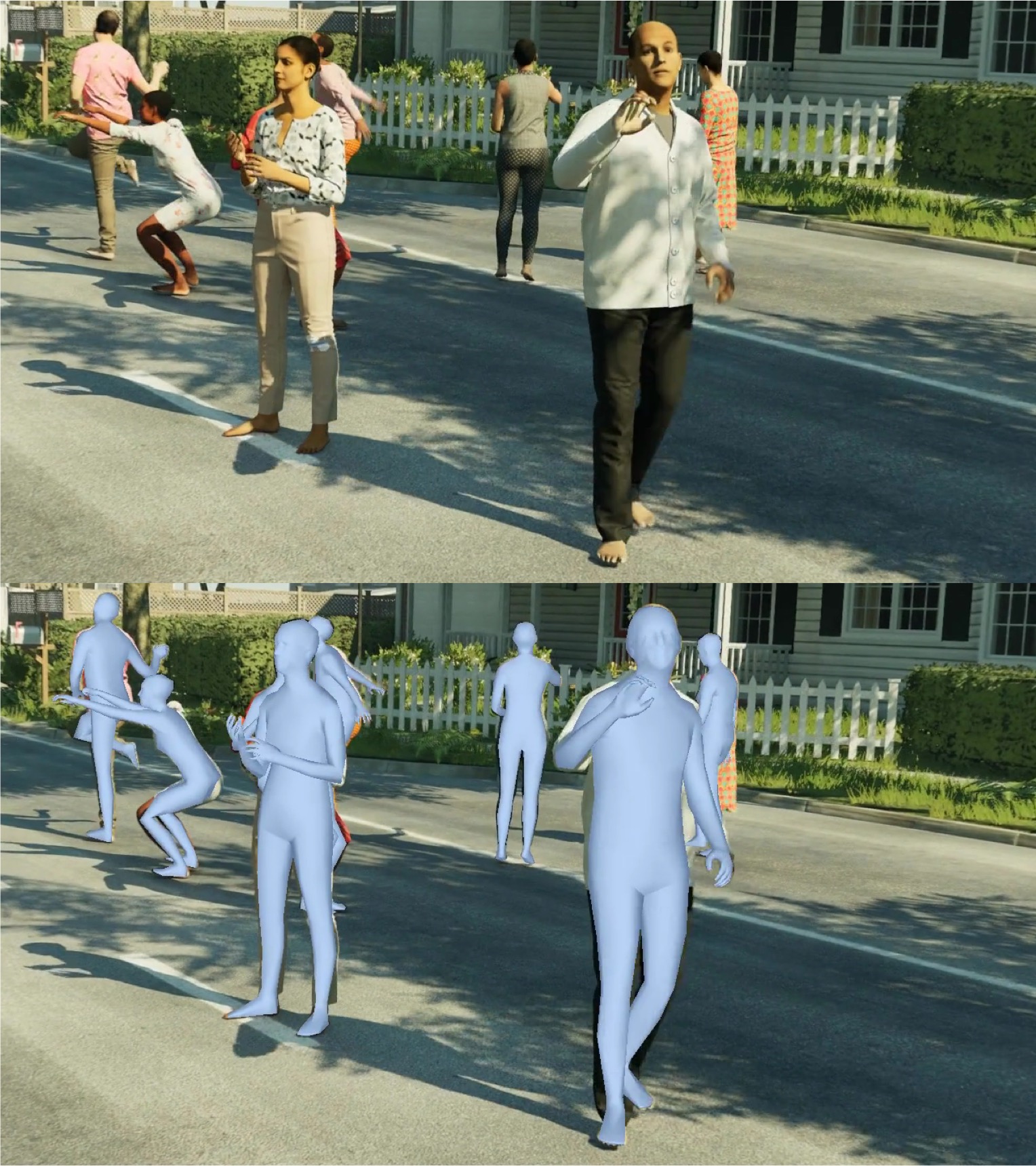}
        \caption{BEDLAM with sparse mesh}
        \label{fig:bedlam_lod}
    \end{subfigure}
    \vspace{-5mm}
    \caption{\textbf{Sparse mesh}. (a) Topology of our sparse mesh, with the red box showing the details of the hands. (b) An example from the BEDLAM~\cite{bedlam} dataset with its annotation converted to the sparse mesh format.}
    \label{fig:sparse_mesh}
\end{figure}

With the regressor matrices, we convert training datasets~\cite{bedlam, amass, 3dpw} to the sparse mesh format. Figure~\ref{fig:bedlam_lod} visualizes an example from the BEDLAM dataset~\cite{bedlam} with its annotation converted to our sparse mesh format. 

\subsection{Sparse mesh to SMPLX}
\label{sec:sparse_to_smpl}
Our sparse mesh can represent detailed motion, but converting it to SMPLX provides a more compact representation and is necessary to compare reconstruction accuracy with prior works. 

The conversion to SMPLX parameters can be achieved through optimization~\cite{smplx, nlf}, but it is slow and prone to local minima. In contrast, we train an optimization-inspired iterative network~\cite{learn2fit, learn2fit2, learn2fit3} for this task. 

\begin{figure}[t!]
    \centering
    \includegraphics[width=0.99\linewidth]{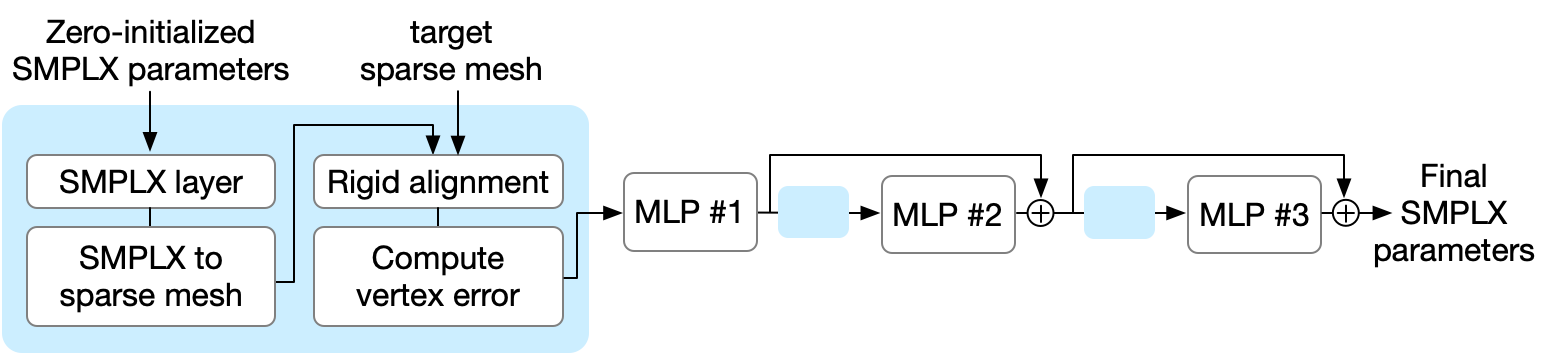}
    \caption{\textbf{Architecture (sparse mesh to SMPLX)}. This network performs iterative refinement to predict SMPLX parameters from a target sparse mesh.}
    \label{fig:sparse_to_smplx}
    \vspace{-2mm}
\end{figure}

Our network performs iterative refinement using a cascade of three MLPs, as shown in Figure~\ref{fig:sparse_to_smplx}. The process begins with zero-initialized SMPLX parameters. At each stage, we generate a predicted sparse mesh from the current SMPLX parameters, using the SMPLX layer followed by our sparse regressor~\ref{sec:smpl_to_sparse}. We then rigidly align the prediction to the target sparse mesh using the head vertices, and compute the per-vertex errors. The MLP takes the errors as input and predicts an update to the parameters. We repeat this process for two more stages. This cascaded design allows each MLP to specialize its refinement.

\begin{figure}[t!]
    \centering
    \includegraphics[width=0.99\linewidth]{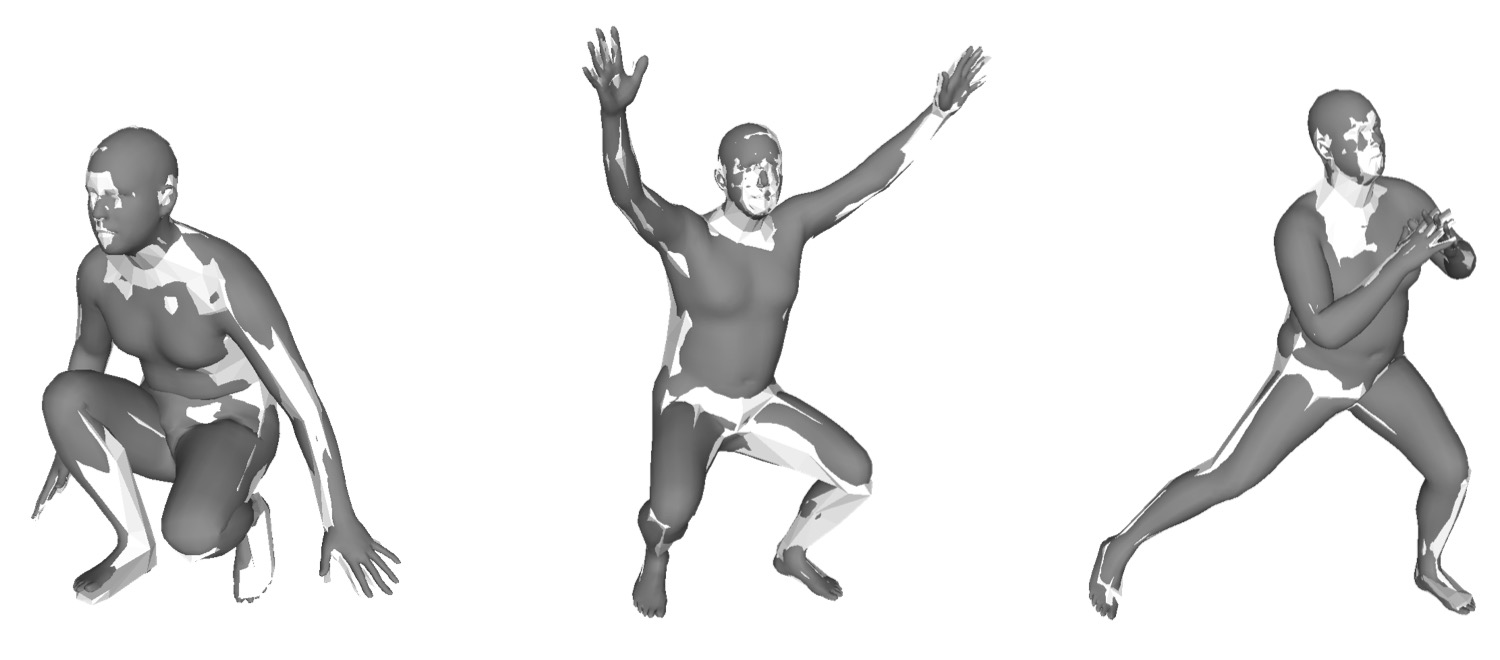}
    \caption{\textbf{Sparse mesh to SMPLX estimation}. Examples of the predicted SMPLX mesh (black) overlaid on the target sparse mesh (white).}
    \label{fig:sparse_to_smplx_ex}
    \vspace{-2mm}
\end{figure}

We tested the network on 3DPW~\cite{3dpw} and find the MPJPE error to be under 5mm. We visualize some qualitative results in Figure~\ref{fig:sparse_to_smplx_ex}.

\section{Dense keypoint detection}
\label{sec:dense_kpt}
The dense 2D keypoints are defined to semantically correspond to the 3D vertices of our sparse mesh. We used the conditional dense body keypoint detection model from SAM 3D Body~\cite{yang2026sam} to infer these 595 body surface keypoints. This model takes a human–centered image and a set of 23 sparse human body keypoints as condition to estimate the dense surface points. We leverage the synthetic dataset BEDLAM~\cite{bedlam} to train the network, where the backbone weights were initialized with ViTPose~\cite{vitpose} to maintain the generalizability to the real images. During inference, we first run ViTPose to extract the sparse 23 keypoints from the input images, which are then fed into the network as the conditioning signal to predict the final 595 dense body surface keypoints. When training the diffusion models, the dense keypoint detection model is frozon.

\begin{figure}[t!]
    \centering
    \includegraphics[width=0.99\linewidth]{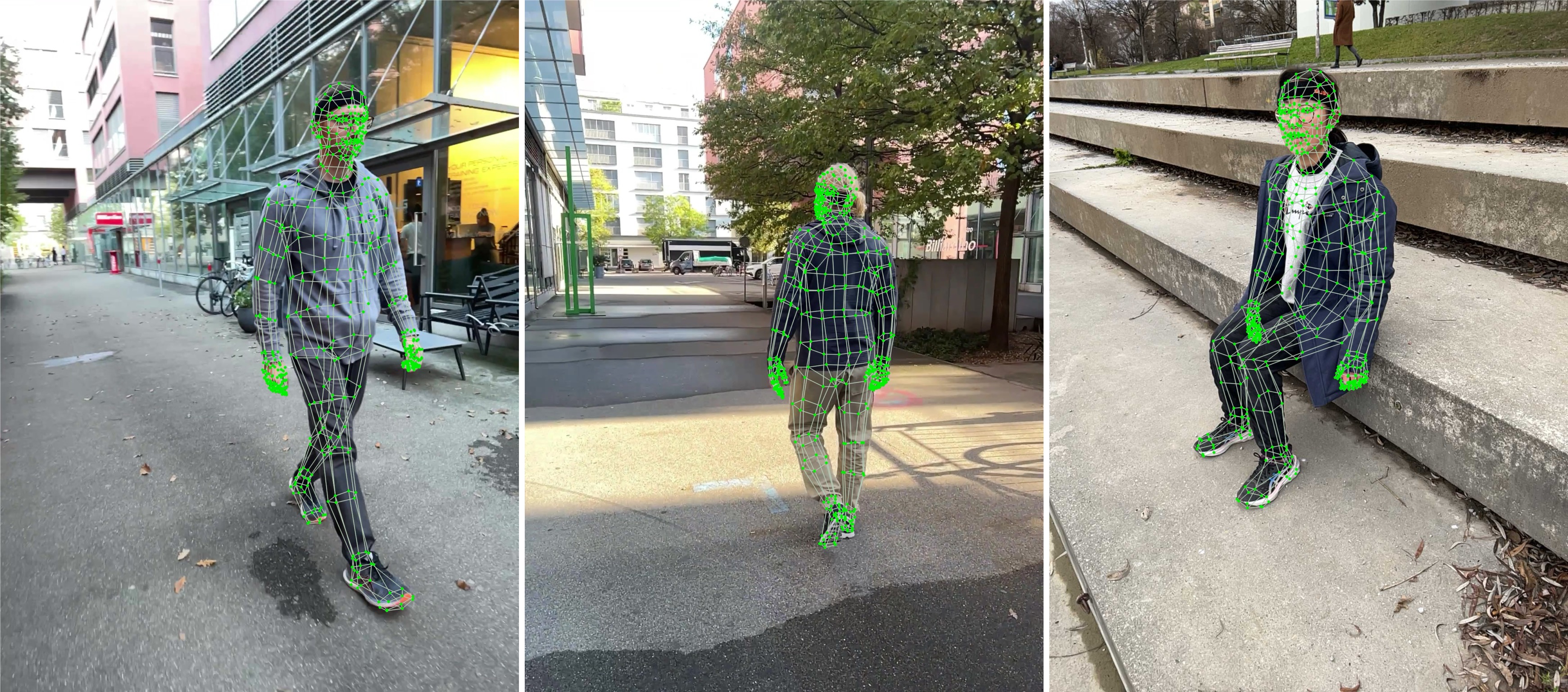}
    \caption{\textbf{Dense keypoint detection} on the EMDB dataset~\cite{emdb}. Edges added for visualization only.}
    \label{fig:dense_kpt}
    \vspace{-3mm}
\end{figure}

Figure~\ref{fig:dense_kpt} visualizes keypoint detection results on the EMDB dataset~\cite{emdb}. Edges from the sparse mesh are added to improve visualization clarity. 

\section{Experiment details}
\label{sec:ex_details}

\subsection{Training}
We train our camera-space model and world-space model with similar setups. We train each model with 8 H200 GPUs, with a batch size of 32 per GPU leading to an effective batch size of 256. The sequence length $T$ is 120 frames.  

\para{Architecture} We use the same diffusion transformer (DiT)~\cite{peebles2023scalable} architecture for the two motion models. It has 8 self-attention layers with $\mathrm{d\_model}=512$, each with 8 heads for multi-head attention, and a hidden dimension of 2048 for the feedforward layer. 

\para{Data} To train our camerea-space motion model, we use AMASS~\cite{amass}, Goliath~\cite{goliath}, BEDLAM~\cite{bedlam} and 3DPW~\cite{3dpw}. For AMASS and Goliath, we generate the 2D dense keypoints from the ground truth annotation and do not use images. For BEDLAM and 3DPW, we generate the 2D dense keypoints and use PromptHMR to extract image features.

To train our world-space motion model, we use AMASS~\cite{amass}, BEDLAM~\cite{bedlam} and 3DPW~\cite{3dpw}. We use the same procedure as training the camera-space model to generate the inputs. We then use the frozen camera-space model and lifting to generate the inputs for the world-space model training. 

We train the dense keypoint detection model with BEDLAM by generating 2D dense keypoints from the ground truth annotation. We train the Sparse-mesh-to-SMPLX iterative network with BEDLAM and 3DPW. 

\para{Augmentation} We employ data augmentation and an augmentation in diffusion noise sampling during training.

We apply augmentation to the generated 2D dense keypoints for training the motion models. We apply noise, perturbation, and masking to the keypoints to simulate detection errors. We apply these augmentation at the point level (e.g. by sampling a set of keypoints) and at the part level (e.g. by applying the same perturbation or masking to all keypoints in a body part).

For training the diffusion models, we apply an augmentation to the sampling of diffusion time step $k$. For 50\% of the samples, we uniformly sample time steps from $\{1, ..., 1000\}$ similar to the standard diffusion training~\cite{ddpm, song2020ddim}. For 50\% of the samples, we set $k=1000$, corresponding to the highest level of corruption. This is equivalent to the ``estimation mode" in GENMO~\cite{li2025genmo}, which finds this strategy encourages the diffusion model to produce good estimation during early diffusion steps. 

\subsection{Evaluation metrics}
\para{Camera-space reconstruction} We evaluate our model's camera-space reconstruction accuracy using three metrics: MPJPE, PA-MPJPE, and PVE. The MPJPE (mean per-joint position error) aligns the prediction with the ground truth at the pelvis location (removing translation) and measures the mean squared error (MSE) on the 3D joints. The PA-MPJPE (Procrustes-aligned MPJPE) rigidly aligns the prediction and ground truth 3D joints (removing rotation, translation and scale) before calculating the MSE on 3D joints. Finally, the PVE (per-vertex error) aligns the predicted and ground truth meshes at the pelvis (removing translation) and computes MSE on the vertices. 

\para{World-space reconstruction} We evaluate world-space reconstruction accuracy with WA-MPJPE, W-MPJPE, RTE, Jitter and Foot sliding. WA-MPJPE and W-MPJPE both measure 3D joints MSE on 100-frame segments of prediction with the ground truth, but differ in how they align the prediction with the ground truth. WA-MPJPE aligns the whole segment (e.g. 3D joints across 100 frames) while W-MPJPE aligns the first two frame (e.g. 3D joints in the first two frames)~\cite{slahmr}. Intuitively, WA-MPJPE measures the accuracy and coherence of the 100-frame motion snippet, while W-MPJPE additionally measures drift. 

RTE (root trajectory error) measures the accuracy of the whole trajectory. It rigidly
aligns the trajectories of the root and computes the mean square error in the unit of \%. Jitter uses finite difference to compute the jerk on the 3D joints to access motion smoothness. Finally, foot sliding calculates the displacement on the predicted foot vertices on contact frames to measure erroneous sliding. 

\begin{figure}[t!]
    \centering
    \includegraphics[width=0.99\linewidth]{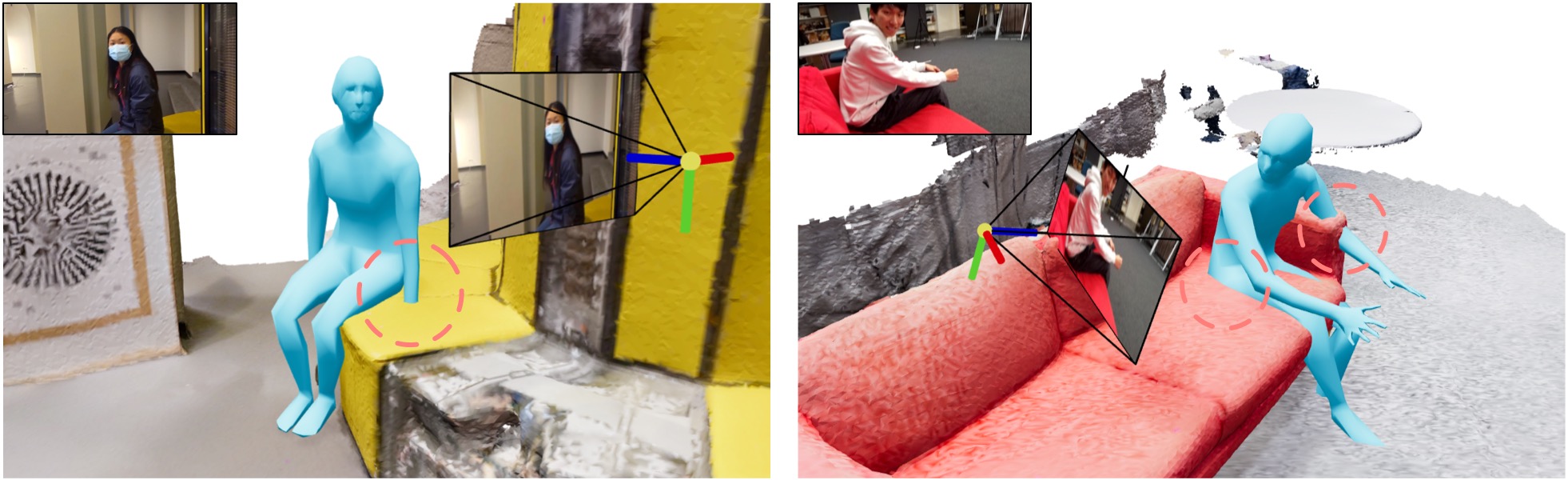}
    \caption{\textbf{Limitation 1}. Because our models do not incorporate 3D scene information, the results exhibit inconsistencies in fine-grained details.}
    \label{fig:limit1}
    \vspace{-2mm}
\end{figure}

\begin{figure}[t!]
    \centering
    \includegraphics[width=0.99\linewidth]{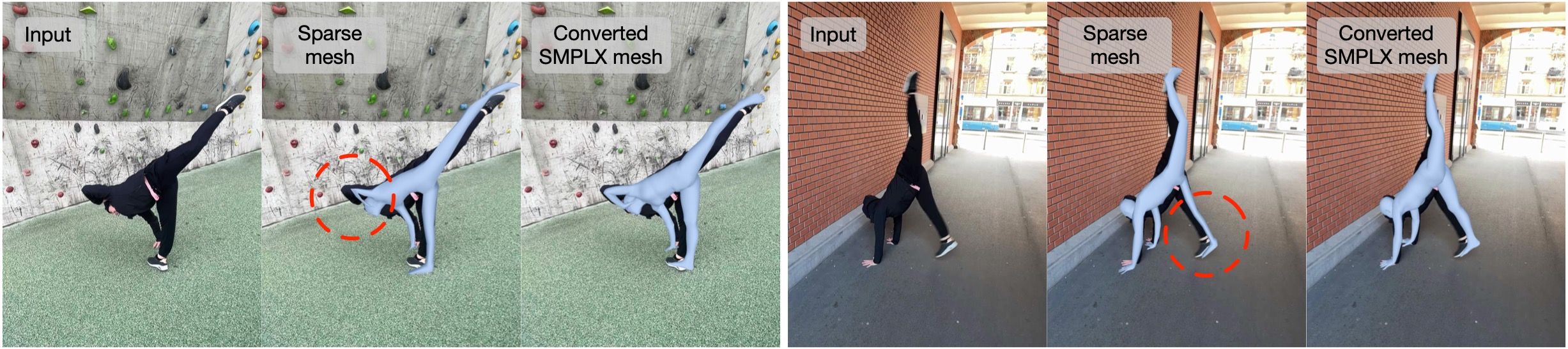}
    \caption{\textbf{Limitation 2}. In challenging poses, the generated sparse meshes sometime exhibit unrealistic deformation. Converting them to SMPL meshes (sec~\ref{sec:sparse_to_smpl}) partially improves the results.}
    \label{fig:limit2}
    \vspace{-2mm}
\end{figure}

\subsection{Limitations}
While DuoMo achieves a new state-of-the-art in world-space human motion reconstruction, it has several limitations that we want to address in future works. 

\para{Scene awareness} While we demonstrated that DuoMo can generate scene-consistent motion with guided sampling in the Experiment section, the 3D scene information is not explicitly used. Figure~\ref{fig:limit1} illustrates some failures in terms of human-scene inconsistency. Future works could incorporate scene information or other physics-based objectives in the guided sampling process~\cite{yuan2023physdiff} to improve such scenarios. Another direction is to train a world-space motion model that can take 3D scene information as auxiliary conditioning. 

\para{Uncertainty awareness} We have modeled visibility at the keypoint level and the frame level, by replacing the embeddings of occluded keypoints or frames with null tokens (Method section). However, the keypoint visibility is determined by thresholding detection confidence, a procedure that is not always accurate. Turning detection confidence into a binary visibility label also discards information. Future work could improve accuracy of the motion model by integrating uncertainty reasoning with detection confidence. 

\para{Mesh generation robustness} We demonstrate that our architecture can generate the motion of mesh vertices, but in difficult cases the generated meshes could have unrealistic deformation. Figure~\ref{fig:limit2} shows examples of deformed meshes. One interesting direction is to integrate our architecture with a tokenized or latent representation of the sparse mesh~\cite{fiche2025mega, lin2025partcrafter} in the form of latent diffusion~\cite{stable_diffusion}.

% WARNING: do not forget to delete the supplementary pages from your submission 
% \input{sec/X_suppl}
% {
%     \small
%     \bibliographystyle{ieeenat_fullname}
%     \bibliography{sec/X_suppl}
% }

\end{document}